# Three Kinds of Negation in Knowledge and Their Mathematical Foundations

**Zhenghua Pan**  Panzh@jiangnan.edu.cn
**Yong Wang**  wymath@jiangnan.edu.cn

*School of Science, Institute of Mathematics and Physics, Jiangnan University, Wuxi 214122, China*

**Abstract**

In the field of artificial intelligence, understanding, distinguishing, expressing, and computing the "negation" in knowledge is a fundamental issue in knowledge processing and research. In this paper, we examine and analyze the understanding and characteristics of "negation" in various fields such as philosophy, logic, and linguistics etc. Based on the distinction between the concepts of "contradiction" and "opposition", we propose that there are three different types of negation in knowledge from a conceptual perspective: "contradictory negation", "opposite negation", and "intermediary negation". To establish a mathematical foundation that fully reflects the intrinsic connections, properties, and laws of these different forms of negation, we introduce "SCOI: sets with contradictory negation, opposite negation and intermediary negation" and "LCOI: logic with contradictory negation, opposite negation and intermediary negation", and we proved the main operational properties of SCOI as well as the formal inference relations in LCOI. Under the three-valued semantic interpretation provided for LCOI, we proved that LCOI is both sound and complete. Furthermore, the objectivity and expressive power of "intermediary negation" in SCOI and LCOI are demonstrated through examples. SCOI and LCOI can express statements of the type '$x$ is neither a positive nor a negative integer' and 'he is neither rich nor poor'. For this type statement with more complex negativity, whether they are clear or fuzzy, they can be expressed by the "intermediary negation" of a set in SCOI and the "intermediary negation" of an atomic formula in LCOI. Finally, to demonstrate that SCOI and LCOI are suitable for the representation and reasoning of knowledge with different forms of negation, we applied SCOI to a multi-attribute decision-making example, showing that both SCOI and LCOI are effective in knowledge processing.

This work was supported by the National Natural Science Foundation of China [60575038, 60973156, 61375004]; the State Key Laboratory for Novel Software Technology, Nanjing University, P.R. China [KFKT2020B01]

# 1. Introduction

"Knowledge" is the mental product of human understanding of the objective world, while concepts are the foundation of thought (Margolis, et al., 2023). From a philosophical perspective, the construction of any knowledge begins with the observation and understanding of things through existing concepts. From the standpoint of artificial intelligence, a fundamental issue in knowledge representation and reasoning is the study and handling of concepts (Ahamed, 2014). The research on knowledge, especially common sense and non-standard knowledge, in artificial intelligence involves distinguishing, expressing, judging, and reasoning about concepts and their relationships, which are fundamental issues (Carter, 2023). "Negation" is an important universal phenomenon in language and a necessary condition for every human language (Horn, et al., 2020). The handling of negation is one of the challenges in natural language processing. With the advent of machine learning and deep learning, the processing of negation has been receiving increasing attention (Makkar, et al., 2023). The understanding and research of negation in knowledge hold the same status and significance as the study of these fundamental issues. Nowadays, there is an increasing amount of research on negation processing across many fields of knowledge, which clearly demonstrates this point.

"Negation" is a phenomenon of semantic opposition. Negation connects one expression with another, the meaning of which is, to some extent, contrary to that of the former. This relationship can be realized in various ways both syntactically and pragmatically, and there are different kinds of semantic opposition (Horn, et al., 2020). Research on the processing of negation has become a hot topic in various fields of study due to its complexity, the multiple linguistic forms it can take, and the different ways it interacts with words within its scope. As a result, the computational processing issues related to negation have not been resolved (Jiménez-Zafra, et al., 2020). For example, in the field of knowledge management, the most advanced CSKBs (commonsense knowledge bases) can generate an uncommon sense knowledge framework that provides information negation about everyday concepts (Arnaout, et al., 2023, 2022). By considering negation and entropy as a new perspective for knowledge management in learning organizations, the study explores their impact on the field of knowledge management (Anjaria, 2020). For the negative sampling techniques, various methods have been proposed, such as discovering high-quality negation signals from implicit feedback (Wang, et al., 2020), directly using cache tracking for negative triples with large gradients (Zhang, et al., 2021), and relationship-enhanced negative sampling for multimodal knowledge graph completion (Xu, et al., 2022). In the field of sentiment analysis, negation mainly comes in two types: implicit and explicit (Makkar, et al., 2023), these negations can be handled using various approaches, such as rule-based methods (Mehrabi, et al., 2015; Haralabopoulos, et al., 2021), dictionary-based methods (Pradhan, et al., 2021; Mukhtar, et al., 2020), as well as machine learning and deep learning methods (Cruz, et al., 2016; Prllochs, et al., 2020; Singh, et al., 2021). In the field of biomedicine, there are two different approaches to handling negation in medical texts: negation detection and negation resolution. Various research methods have been proposed for this purpose (Chen, 2019; Jong, et al., 2022; Fabregat, et al., 2023; Slater, et al., 2021; Zhu, et al., 2023).

These studies on negation handling in different fields have a cognitive characteristic regarding "negation", which is that there is only one type of negation in the knowledge. In other words, these studies are based on classical logic.

With the development of artificial intelligence, there are new demands for the understanding and handling of "negation" in knowledge. Some studies advocate for the distinction of different types of negation within knowledge (Yang, et al., 2023; Wagner, 2003; Analyti, et al., 2004; Dimiter, 2005; Beeson, et al., 2005; Dung, et al., 2002; Kaneiwa, 2007; Ferré, 2006). In these studies, Wagner (2003) and Analyti et al. (2004) believe that in any information computing system, the concept of negation plays a special role, and that negative information holds equal value to positive information. They propose the distinction between 'strong negation' and 'weak negation' in information computation systems, where strong negation indicates *explicit falsity*, and weak negation represents *non-truth*. Kaneiwa (2007) argues that Description Logic should distinguish between these two types of negation, while Ferré (2006) suggests differentiating 'negation', 'opposition' and 'possibility' in Logical Concept Analysis. These studies on different types of negation in information have a particular characteristic: they mainly focus on the practical needs of knowledge processing, proposing methods to distinguish and handle different negations in knowledge from a semantic perspective. However, they do not recognize or differentiate the various types of "negation" in knowledge at a conceptual level, nor do they explore the theoretical foundations that could reflect the nature, relationships, and laws of these different types of negation (Pan, 2010, 2017).

In our past research (Pan, et al., 2010, 2011, 2013, 2017), we primarily focused on the issue of "negation" in fuzzy knowledge, proposing that three types of negation should be distinguished in fuzzy knowledge, and constructed a fuzzy set and fuzzy logic that incorporate these three types of negation. Building upon this foundation, the present paper expands the scope of research to propose that there are three different types of negation present in general knowledge (including theoretical knowledge, common knowledge, experiential knowledge, etc.) and establishes a mathematical foundation (sets and logic) that can fully reflects the intrinsic connections, properties, and laws of these different forms of negation.

The main contributions of this paper are summarized as follows:

- We have conducted an examination and analysis of various understandings and characteristics of "negation" in different knowledge fields. Based on this, and distinguishing between the concepts of "contradiction" and "opposition", we propose that there are three different types of negation in knowledge: 'contradictory negation', 'opposite negation' and 'intermediary negation'.
- We propose a "S$_{COI}$: sets with contradictory negation, opposite negation, and intermediary negation" and a "L$_{COI}$: logic with contradictory negation, opposite negation, and intermediary negation". Furthermore, we discuss the main operational properties of S$_{COI}$, the formal inference relations in L$_{COI}$, and the semantics of L$_{COI}$.
- The intermediary negation in S$_{COI}$ and L$_{COI}$ can express statements like '$x$ is neither a positive nor a negative integer' and 'he is neither rich nor poor'. In other words, this type of statement can be represented by the intermediary negation of a set in S$_{COI}$ and the intermediary negation of an atomic formula in L$_{COI}$. (If this type of statement is formally expressed using other methods, the expression would become more complicated due to the inability to distinguish the three different types of negation and their relationships that exist within the concepts.)
- In order to demonstrate that S$_{COI}$ and L$_{COI}$ are suitable for handling knowledge and its various forms of negation, we apply S$_{COI}$ and L$_{COI}$ to a multi-attribute decision-making example, showing that they are effective in knowledge processing.

This paper is organized as follows. Section 2 examines and analyzes various understandings and characteristics of "negation" in philosophy, logic, linguistics and other fields. In Section 3, based on the distinction between "clear knowledge" and "fuzzy knowledge", as well as the distinction between the concepts of "contradiction" and "opposition", we propose that there are three different negations between species under a genus concept. In Section 4, we introduce a "set S$_{COI}$ and logic L$_{COI}$ with contradictory negation, oppositional negation and intermediary negation" and study their main properties and characteristics, demonstrating the objectivity and expressive power of "intermediary negation" in S$_{COI}$ and L$_{COI}$ through examples. In Section 5, we explore the application of S$_{COI}$ and L$_{COI}$ in a typical knowledge processing instance. Section 6 presents the conclusions and prospects for future work.

## 2. The Cognition and characteristics of "negation" across different knowledge fields

Over the past 25 centuries, descriptive, theoretical, and empirical studies on "negation" have largely focused on the relatively apparent, complex, or subjective nature of negative statements (Speranza, et al., 2010; Horn, 2004). However, there have been differences in understanding the concept of "negation", and a unified recognition has yet to be established. Therefore, it is necessary to examine and analyze the cognition and characteristics of the concept of "negation" in fields such as philosophy, logic, linguistics, and other knowledge fields.

**I.** In the philosophy, from ancient times to the twentieth century, philosophers have consistently believed that knowing "negation" is more difficult than knowing "affirmation" (Turri, 2022). From the philosophy of Kant, Fichte, and Schelling, which represents German classical philosophy, to Hegel's philosophy, there has never been a unified understanding of the concept of "negation" (Smith, 2022). Since ancient Greece, philosophers have been striving to find the source of negation in the empirical world (Ayer, 1952). Firstly, if the concept of "negation" is treated as a logical category, abstracting all things in the world into various concepts, and then inferring their necessity from the logic between concepts, the law of change between things can thus be grasped, thereby attaining our rational "knowledge". To acquire this kind of knowledge of regularity, formal logic is an important tool. This logic requires sensory experience as the material to fill its "content". However, traditional European philosophy, in the conventional sense, is precisely not satisfied with merely filling the content of experience; it seeks to question the 'foundation' and 'source' of this formal logic. Starting with Kant, German classical philosophers attempted to find a deeper foundation for this formal logic, so as to ground formal logic upon that foundation (Ayer, 1952). To this end, Hegel strengthened the role of "negation," introducing negation into the logical category of ontology, thereby completing the transformation of traditional formal logic by German classical philosophy (Frege, 1919). It can be said that the concept of negation is the core concept that distinguishes Hegel's philosophy from other German classical philosophies, and the entire structure of Hegel's philosophical system is founded upon it (Wretzel, 2014).

For the concept of "negation", Hegel demonstrates that all "negation" is determination. This determined "negation" acquires new content and becomes a new concept. Because it serves as the negation or antithesis of the preceding concept, it is higher and richer than the preceding concept (Gale, 1976). Hence, it is evident that the "negation" Hegel refers to is the negation of itself. Negation is premised on affirmation (Deng, 2022). Affirmative propositions precede negative propositions and are more readily known than negative propositions because affirmation explains negation, just as existence precedes non-existence (Horn, et al., 2020).

**II**. In the classical logic and non-classical logic, there are essential differences in the understanding and recognition of the concept of "negation". In classical logic, due to the adoption of bivalent semantics, the law of excluded middle holds universal applicability. The bivalent semantics of classical logic asserts that a proposition is either true or false, with truth and falsehood mutually negating each other. The concept of negation in classical logic is based on the two-valued semantics of static cognition of things, and has the feature of 'either this or that', so it has its inherent cognitive defects. This is primarily reflected in the difficulty of grasping and characterizing the negation relationship between thing in processes of change, transition, and development, where the original concept of negation becomes inadequate. The evolving understanding and transformation of the concept of "negation" was one of the primary intellectual motivations for the emergence and development of non-classical logic in the early 20th century (Malinowski, 2007).

In the understanding and handling of the concept of "negation" in non-classical logic, Three-Valued Logic is representative. The fundamental idea of three-valued logic is that propositions in a propositional calculus system can take not only the traditional binary values of true and false but also an intermediate or modal value (Malinowski, 2007). In three-valued logic, the negation of true and false remains consistent with traditional logic, with the only issue to be addressed being how to define the "negation" of the third value. Regarding this, Łukasiewicz and Post defined the negation of the third value as undecided or possible. The concept of "negation" in Quantum Logic is the most intriguing aspect of non-classical logic. To address the problem that "quantum probability propositions in the double-slit experiment do not adhere to the law of excluded middle in classical logic", Reichenbach proposed quantum logic with three truth values: true, false, and indeterminate (Horn, 1989).

**III**. In the linguistics, the concept of "negation" is a complex phenomenon in both logic and natural language (Horn, et al., 2020). Negation is not only an operation in formal logic but also an operation in natural language, where by adding 'not' or other negation cues; a proposition is replaced by its contradictory proposition. In natural language, the meaning of negation is often less explicit and consists of different components: there is no precisely defined negation operator but rather a negation cue, which may be a syntactically independent negation marker such as 'never', 'nor' and 'not' (syntactic negation); it may also be an affix indicating negation, such as 'i(n)-' and 'un-' (morphological negation); or it may be a word containing a negation component, such as 'deny' (lexical negation). The scope of negation refers to the entire portion of a sentence or discourse that is negated, i.e., the part affected by the negation cue (Schon. et al., 2021). In understanding the concept of "negation" from the perspectives of language form and semantics, the famous book 'A Natural History of Negation' (Horn, 1989) suggests that to avoid the assumed need for introducing an appropriate semantic operator, it is recommended to expand the metalinguistic usage of negation.

**IV**. In the applied logic, the nature of negation combined with the characteristics of other logical operations and the structural features of deductive relations, serves as a bridge between different logical systems. Therefore, negation plays an important role when selecting a logical system for specific applications (Dov, et al., 1999). In this monograph by Dov, et al.(1999), some scholars have proposed different perspectives on the concept of "negation" in the field of applied logic. Among them, one argues that negation should be considered from two distinct perspectives—syntactic and semantic—thereby identifying two different types of negation; one suggests that the current orthodox view of logic has gone astray regarding negation, the negation is an entity, an operation that is both one (a determinable, though widely used but far from orthodox, operation) and many (having multiple determinate entities); one contends that negation should be viewed as primitive, and one should explain how the understanding of its meaning arises from the interpretation of primitive metaphysical opposites in the language of interpretation.

In addition to the aforementioned differing views on the concept of "negation", new mathematical theories have been explored through the study of the concept of negation. Among these, Kwuida (2004) aimed to extend FCA (Formal Concept Analysis) to a broader domain by introducing the concept of "negation" of formal concepts. To achieve this, the concepts of 'weak negation' and 'weak opposition' were proposed, which introduced an internal negation operation into concept algebra, thereby extending Boolean algebra. Almeida (2009) investigated lattices with negation operations and their properties and structures. Through the study of different negation operations, new perspectives and theoretical foundations have been provided for the canonical extension and relational representation of lattices.

Through the above investigation and analysis of the cognition and understanding of the concept of "negation" in philosophy, logic, linguistics, and applied logic, we conclude that these cognitions and understandings are not entirely consistent and emphasize different aspects. However, it can be observed that they exhibit the following fundamental characteristics:

- The basic characteristic of the "negation" of a concept is the negation of itself.
- The concept of "negation" is relative to a specific concept ("affirmative concept"). The concept of "negation" takes to negate the intension of "affirmative concept" as its own intension, and takes the scope of objects outside the extension of "affirmative concept" as its own extension.
- Logically, "negation" presupposes "affirmation", and affirmative propositions precede negative propositions.

## 3. Three kinds of "negation" in knowledge

Based on the above characteristics of the cognition and understanding of the concept of "negation" in philosophical and other knowledge domains, this section distinguishes between "clear knowledge" and "fuzzy knowledge" at the conceptual level, fully understanding the "contradiction" and "opposition" within concepts, thereby proposing that there are three different types of negation in knowledge and discussing their characteristics.

**3.1 Distinguishing "clear knowledge" and "fuzzy knowledge"**

The term "concept" has two basic attributes: the intension of the concept and the extension of the concept. The intension of concept refers to the meaning of concept; the extension of concept refers to the range of object that the concept reflects. The logical method of determining a concept is to determine the connotation and extension of the concept, so that the category of the concept can be determined (Kwiatkowski, 2006). Some philosophers maintain that possession of natural language is necessary for having any concepts (Margolis, et al., 2023). According to this philosophical view, the problem of non-precise or vague knowledge, its representation and conceptualization already has a long tradition and is still regarded as one of the most worthy of discussion. It grew out of the philosophical reflection on vague concepts of colloquial language and on the value of this language for philosophy and science (Wybraniec-Skardowska, 2001). Thus, for a concept, we can divide it into "clear concept" or "fuzzy concept" based on its intension and extension.

A concept is called "clear concept" if the intension is certain and the extension is clear. A concept is called "fuzzy concept" if the intension is certain and the extension is vague.
In short, whether a concept is clear or fuzzy is determined by whether its extension is clear or fuzzy.

For example, under the genus concept "number", the species concepts 'rational number' and 'irrational number' are clear concepts because their extensions are clear. Under the genus concept "person", the species concepts 'young person' and 'non-young person' are fuzzy concepts because their extensions are vague.

We apply the above idea of distinguishing concepts to the classification of knowledge, thereby distinguishing knowledge (theoretical knowledge, common sense, and empirical knowledge) into *clear knowledge* and *fuzzy knowledge*.

In knowledge statements, if all the concepts contained in the statement are clear concepts, the knowledge is classified as *clear knowledge*. For example, professional knowledge: "The set of real numbers consists of rational and irrational numbers". Since the concepts "real number", "rational number", and "irrational number" are all clear concepts, this knowledge is classified as clear knowledge. Common sense knowledge: "China has an area of 9.6 million square kilometers." Since the concepts "China", "9.6 million square kilometers", and "area" are all clear concepts, this knowledge is also classified as clear knowledge.

In knowledge statements, if there are fuzzy concepts within the concepts contained in the statement, the knowledge is classified as *fuzzy knowledge*. For example, the knowledge statement "Zhang San is a top student at Peking University" is fuzzy knowledge because the concept "top student" is a fuzzy concept. Similarly, the statement "If the motor temperature is high and the operating time is long, the machine will shut down" is fuzzy knowledge, as the concepts "high temperature" and "long time" are fuzzy concepts.

**3.2 Distinguishing "contradiction" and "opposition" in concepts**

In various concepts that reflect the essence of things, the distinction between the concepts of "contradiction" and "opposition" has been made since Aristotle's logic (Leszl, 2004). Regarding the concepts of "contradiction" and "opposition," there is now a unified understanding in formal logic (Kwiatkowski, 2006; Horn, 2018): (1) The relationship between concepts is not the relationship inherent in the objects reflected by the concepts but rather the relationship between the extensions of the concepts. (2) A concept (affirmative concept) and its "negation" are two different species concepts under the same genus concept. (3) "Contradiction" and "opposition" are two concepts of fundamentally different nature. A contradictory concept refers to two different species concepts under the same genus concept that have a contradictory relationship, where their intensions negate each other, their extensions are mutually exclusive (either-or), and the sum of their extensions equals the extension of the genus concept. An oppositional concept, on the other hand, refers to two different species concepts under the same genus concept that have an oppositional relationship, where their intensions negate each other and exhibit the greatest difference, their extensions mutually exclude each other, and the sum of their extensions is less than the extension of the genus concept.

The concept of "negation" is usually semantically limited to contradictions and oppositions between propositions (Horn, et al., 2020). Therefore, we naturally ask: For a species concept under a genus concept, is the concept that has a contradictory relationship with it its "negation"? Is the concept that has an oppositional relationship with it also its "negation"? Beyond this, are there other forms of "negation"? To address these questions, we need to examine and analyze the essence of "contradiction" and "opposition" from a conceptual perspective.

Hegel was able to establish logic as a science due to his understanding and recognition of the concept of "negativity." According to Hegel, the process of the emergence and development of contradiction begins as the unity of different determinations, then produces difference, turns into opposition, and finally manifests as contradiction. In opposition, the differing entity is not just any other entity but rather the one that is directly opposite to it. Hegel's emphasis on the differences and oppositions of things was aimed at further considering and advancing another kind of determination. This determination distinguishes a form of negation that negates itself and reflects itself in opposition, which is "contradiction" (Wretzel, 2014).

The aforementioned dialectical analysis by Hegel, on the level of foundational thought, distinguishes between the "oppositional" and "contradictory" things and their relationships, differentiating the concepts of "opposition" and "contradiction" within the realm of abstract concepts. The important distinctions between "contradiction" and "opposition" in the context of Hegel's logic are as follows (Malpass, et al., 2017; Sethy, 2021):

- "Contradiction is the unity of opposites" defines the relationship between 'contradiction' and 'opposition', but it also reveals that 'contradiction' and 'opposition' are distinct concepts.
- "Contradiction" is an abstract concept that reflects the mutual negation of "what something is" and "what it is not" during its process of change. It is a form of self-negation of a thing in its change. "Opposition," however, reflects the mutual negation of 'two things' both in their change and in their constancy. It is the mutual negation of two things.
- "Contradiction" and "opposition" are different forms of negation, with "opposition" being the extreme form of "contradiction".

In fact, Hegel's term "contradiction" can only be applied in an abstract, conceptual context and not to specific things [46]. For example, the contradictory counterpart of "man" is "non-man". There is no such thing as a "non-man" in the objective world; it can only be an abstract concept that encompasses numerous entities. However, the concept of "opposition" is essentially different from "contradiction." In affirming itself, "opposition" also affirms the existence of the entity opposite to itself. For instance, "daytime" is the opposite of "night," and "daytime" is a phenomenon that exists in the objective world. Moreover, there are other objective phenomena that exist between "daytime" and "night," which serve as the "intermediary" between the two.

From Hegel's philosophy to dialectical materialism, it is believed that all oppositions transition through "intermediaries" (intermediate links) (Horn, 2010). According to Hegel, a pair of opposing concepts flows between two extremes through "intermediaries". An intermediary is the link between opposing categories, and such intermediary links can be found in everything, everywhere, and in every concept. Engels, in Dialectics of Nature, systematically and thoroughly elaborated on the objectivity and universality of "intermediary" between opposites, emphasizing that all opposites transition through "intermediary" (Kangal, 2019).

In summary, the following two conclusions can be drawn:
a. In concepts, "contradiction" and "opposition" should be distinguished. Contradiction is an abstract concept, a form of self-negation of a thing. Opposition is a concept of mutual negation between two things. "Contradiction" and "opposition" are two different forms of negation, with opposition being the extreme form of contradiction".
b. The "mediation" that exists between opposing concepts (referred to as "intermediary concept" for short) serves as a bridge connecting the two extremes of opposition. There is no opposition without "intermediary", and all opposite concepts transition through "intermediary".

## 3.3 Three kinds of negation in concepts and their characteristics

Based on the above examination and analysis of the concepts of "contradiction" and "opposition" from the essence of concepts, we propose that there are three different forms of "negation" in both clear and fuzzy concepts:

(1) *Contradictory Negation*. For a species concept under a genus concept, another species concept that has a contradictory relationship with it constitutes a form of negation. We refer to this type of negation as "contradictory negation". In this form of negation, the intensions (connotations) of the two species concepts mutually negate each other, the extensions (denotations) are mutually exclusive (either one or the other), and the sum of the extensions equals the extension of the genus concept.

(2) *Opposite Negation*. For a species concept under a genus concept, another species concept that has an oppositional relationship with it constitutes another form of negation. We refer to this type of negation as "opposite negation". In this form of negation, the intensions of the two species concepts mutually negate each other and exhibit the greatest difference in intension, but their extensions are not mutually exclusive (not either-or), and the sum of their extensions is less than the extension of the genus concept.

(3) *Intermediary Negation*. The intermediary concept between opposite concepts constitutes a (weak) form of negation of the opposite concepts. We refer to this type of negation as "intermediary negation". In this form of negation, the opposite concepts transition through the intermediary concept and the sum of their extensions equals the extension of the genus concept.

We must point out that in reality there are cases where "contradiction" and "opposition" appear identical. Such situations should be understood as contradiction rather than opposition. For example, under the genus concept of 'real numbers', the species concepts 'rational numbers' and 'irrational numbers' (i.e., non-rational numbers) are both contradictory and oppositional. However, since there is no "intermediary" between rational and irrational numbers, they are not oppositional concepts.

To fully understand the implications of the three types of negation described above, we will further discuss their characteristics in terms of the relationships between intension and extension in concepts.

**(1) Contradictory Negation in Clear Concepts (CNC)**

Characteristics of CNC: Extensions are clear, either this or that, and the sum of extensions is equal to the extension of the genus concept.

For example, the positive integer and non-positive integer under the genus concept of "integer" are clear concepts, while the non-positive integer is the contradictory negation of positive integer. The diagram illustrating the extensional relationship between them is shown below (Figure 1).

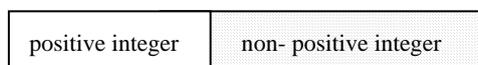

**Fig. 1**    The extensional relationship between positive integer and non-positive integer

**(2) Opposite Negation in Clear Concepts (ONC)**

Characteristics of ONC: Extensions are clear, not "either this or that", and the sum of the extensions is less than the extension of the genus concept.

For example, the positive integer and negative integer under the genus concept of "integer" are clear concepts, while the negative integer is the opposite negation of positive integer. The diagram illustrating the extensional relationship between them is shown below (Figure 2).

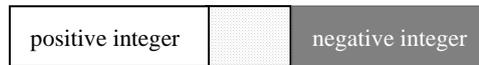

**Fig. 2**　The extensional relationship between positive integer and negative integer

### (3) Intermediary Negation in Clear Concepts (INC)

Characteristics of INC: Extensions are clear, opposing sides transition to each other through 'intermediaries', and the sum of the extensions equals the extension of the genus concept.

For example, zero, positive integer and negative integer are clear concepts under the genus concept of "integer". Zero serves as an 'intermediary' between positive integer and negative integer, and it is the intermediary negation of both positive integer and negative integer. The diagram illustrating the extensional relationship between them is shown below (Figure 3).

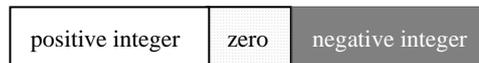

**Fig. 3**　The extensional relationship between positive integers, negative integers, and zero.

### (4) Contradictory Negation in Fuzzy Concepts (CNF)

Characteristics of CNF: Extensions are not clear, either this or that, and the sum of extensions is equal to the extension of the genus concept.

For example, the daytime and non-daytime under the genus concept of "day" are fuzzy concepts, while the non-daytime is the contradictory negation of daytime. The diagram illustrating the extensional relationship between them is shown below (Figure 4).

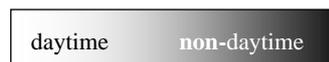

**Fig. 4**　The extensional relationship between daytime and non-daytime

### (5) Opposite Negation in Fuzzy Concepts (ONF)

Characteristics of ONF: Extensions are not clear, not "either this or that", and the sum of the extensions is less than the extension of the genus concept.

For example, the daytime and night under the genus concept of "day" are fuzzy concepts, while the night is the opposite negation of daytime. The diagram illustrating the extensional relationship between them is shown below (Figure 5).

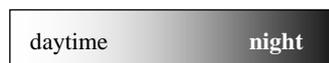

**Fig. 5**　The extensional relationship between daytime and night

### (6) Intermediary Negation in Fuzzy Concepts (INF)

Characteristics of IFC: Extensions are not clear, opposing sides transition to each other through 'intermediaries', and the sum of the extensions equals the extension of the genus concept.

For example, dusk, daytime and night are fuzzy concepts under the genus concept of "day". Dusk serves as an 'intermediary' between daytime and night, and it is the intermediary negation of both daytime and night. The diagram illustrating the extensional relationship between them is shown below (Figure 6).

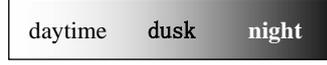

**Fig. 6** The extensional relationship between daytime, night, and dusk

It is evident that, because concepts are the most fundamental components of knowledge, these three different negations also exist within knowledge itself.

## 4. Set and logic with three kinds of negation

In this section, we introduce and discuss the mathematical foundations (set theory and logic) for the three different negations in knowledge, based on our previous research (Pan, et al., 2010, 2011, 2013, 2017). One is referred to as "SCOI: Set with contradictory negation, opposite negation, and intermediary negation"; the other is referred to as "LCOI: Logic with contradictory negation, opposite negation, and intermediary negation". We also discuss the operations and properties of SCOI, as well as the formal reasoning relation in LCOI and the semantics of LCOI.

### 4.1 SCOI: Sets with contradictory negation, opposite negation and intermediary negation

In this section, we use symbols ¬, ⇁, and ~ to represent 'contradictory negation', 'opposite negation', and 'intermediary negation', respectively. And, continue to use the definition of fuzzy sets (Zadeh, 1965).

**Definition 1**. Let $U$ be universe of discourse, $\lambda \in (0, 1)$. Mapping $f: U \to [0, 1]$ confirms a set $A$ on $U$, call $f$ the membership function of $A$, and $f(x)$ the membership degree of $x$ to $A$ (denoted as $A(x)$).

(1) If $A$ is a fuzzy set, then

(i). Mapping $f^⇁: \{A(x) \mid x \in U\} \to [0, 1]$ confirms a fuzzy set $A^⇁$ on $U$, $A^⇁(x) = f^⇁(A(x)) = 1 - A(x)$. Call $A^⇁$ the opposite negation set of $A$.

(ii). Mapping $f^\sim: \{A(x) \mid x \in U\} \to [0, 1]$ confirms a fuzzy set $A^\sim$ on $U$, $A^\sim(x) = f^\sim(A(x))$. Call $A^\sim$ the intermediary negation set of $A$. Where

$$A^\sim(x) = \begin{cases} \lambda - \dfrac{2\lambda-1}{1-\lambda}(A(x)-\lambda), & \text{when } \lambda \in [½, 1) \text{ and } A(x) \in (\lambda, 1] \quad (a) \\[6pt] \lambda - \dfrac{2\lambda-1}{1-\lambda}A(x), & \text{when } \lambda \in [½, 1) \text{ and } A(x) \in [0, 1-\lambda) \quad (b) \\[6pt] 1 - \dfrac{1-2\lambda}{\lambda}A(x) - \lambda, & \text{when } \lambda \in (0, ½] \text{ and } A(x) \in [0, \lambda) \quad (c) \\[6pt] 1 - \dfrac{1-2\lambda}{\lambda}(A(x)+\lambda-1) - \lambda, & \text{when } \lambda \in (0, ½] \text{ and } A(x) \in (1-\lambda, 1] \quad (d) \\[6pt] A(x), & \text{other} \quad (e) \end{cases}$$

(iii). Mapping $f^¬: \{A(x) \mid x \in U\} \to [0, 1]$ confirms a fuzzy set $A^¬$ on $U$, $A^¬(x) = f^¬(A(x)) = max(A^⇁(x), A^\sim(x))$. Call $A^¬$ the contradictory negation set of $A$.

(2) If $A$ is a clear set, then $A(x) \in \{0, 1\}$, $A^⇁(x) = 1 - A(x)$, $A^\sim(x) = ½$, $A^¬(x) = max(A^⇁(x), A^\sim(x))$.

The set on the domain $U$ determined above is called "Sets with contradictory negation, opposite negation and intermediary negation", for short SCOI. The set of SCOI on $U$ is denoted as $SCOI(U)$.

For any $A \in SCOI(U)$, according to Definition 1, $A$ and its contradictory negation set $A^¬$, opposite negation set $A^⇁$ and intermediate negation set $A^\sim$ have following characteristics:

- $\forall x \in U. A(x), A^⇁(x), A^\sim(x), A^¬(x) \in [0, 1]$.
- $\lambda$ is a variable parameter. The size and variation of $\lambda$ determine the size and range of values for $A(x)$, $A^⇁(x)$ and $A^\sim(x)$, i.e. $\lambda$ is a "*threshold*" for the range of the values of these membership degrees.

- $A$ is a clear set, i.e. the special case of the fuzzy set $A$ when $A(x) \in \{0, 1\}$ and $\lambda = \frac{1}{2}$.

In Definition 1, how to determine the membership degree $A^\sim(x)$ of x to fuzzy set $A^\sim$, the expression (a)-(e) is the key to the definition. The basic idea is as follows:

Since $A(x), A^\daleth(x), A^\sim(x), A^\neg(x) \in [0, 1]$, in order to determine their value range in $[0, 1]$, we introduce a parameter variable $\lambda$ in the open interval $(0, 1)$. Thus, when $\lambda \geq \frac{1}{2}$, $[0, 1]$ is divided into three subintervals $[0, 1-\lambda)$, $[1-\lambda, \lambda]$ and $(\lambda, 1]$. If $A(x) \in (\lambda, 1]$, then $A^\daleth(x) \in [0, 1-\lambda)$ by (i). At this point, if $A^\sim(x) \in [1-\lambda, \lambda]$, since there is a one-to-one correspondence in between the values of AA and BB, the expression (a) can be obtained according to the principle that the points in the pairwise disjoint interval in the function of real variable have a one-to-one correspondence. If $A(x) \in [0, 1-\lambda)$ and $A^\sim(x) \in [1-\lambda, \lambda]$, we can get the expression (b) for the same reason. When $\lambda \leq \frac{1}{2}$, $[0, 1]$ is divided into three subintervals $[0, \lambda)$, $[\lambda, 1-\lambda]$ and $(1-\lambda, 1]$. Similarly, expressions (c) and (d) can be established. In other cases, $A^\sim(x) = A(x)$, i.e. (e).

We can use the following graph (Fig. 7) to represent the interrelationships between $A(x)$, $A^\daleth(x)$ and $A^\sim(x)$ (the symbols "•" and "○" in the figure represent the close endpoint and the open endpoint of an interval, respectively).

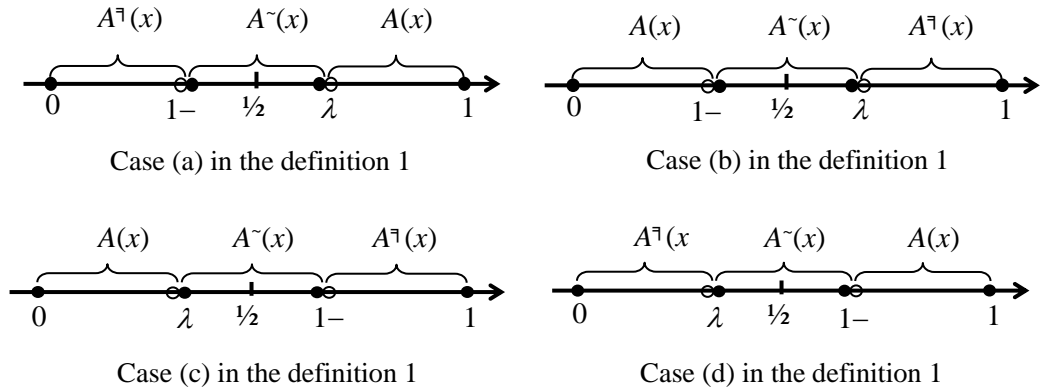

Case (a) in the definition 1    Case (b) in the definition 1

Case (c) in the definition 1    Case (d) in the definition 1

**Fig. 7**   The interrelationships between $A(x)$, $A^\daleth(x)$ and $A^\sim(x)$

### 4.1.1 The Operations and Properties of SCOI

According to Definition 1, the following properties of SCOI can be proven.

**Proposition 1**. $\forall A \in SCOI(U)$. Then
- $A(x) \geq A^\sim(x) \geq A^\daleth(x)$, or $A^\daleth(x) \geq A^\sim(x) \geq A(x)$.
- $A(x) > A^\sim(x) > A^\daleth(x)$, if and only if $A(x) > \frac{1}{2}$.
- $A(x) < A^\sim(x) < A^\daleth(x)$, if and only if $A(x) < \frac{1}{2}$.
- $A(x) = A^\sim(x) = A^\daleth(x)$, if and only if $A(x) = \frac{1}{2}$.

**Proposition 2**. $\forall A \in SCOI(U)$. Then
- $A^\neg(x) = A^\daleth(x)$, when $A(x) \leq \frac{1}{2}$.
- $A^\neg(x) = A^\sim(x)$, when $A(x) \geq \frac{1}{2}$.

In SCOI, $\forall a, b \in [0, 1]$, if $a \leq b$ then $f^\daleth(a) \geq f^\daleth(b)$, $f^\sim(a) \geq f^\sim(b)$. Therefore, the following proposition 3 holds.

**Proposition 3**. In SCOI, $f^\daleth$ and $f^\sim$ are both decreasing functions.

In SCOI, the operations of equality, inclusion, union and intersection of sets maintain the definitions and properties of the basic operations of classical sets and fuzzy sets.

**Definition 2**. $\forall A, B \in SCOI(U)$. For all $x \in U$,
$$A = B, \text{ if and only if } A(x) = B(x);$$
$$A \subseteq B, \text{ if and only if } A(x) \leq B(x).$$

**Definition 3**. $\forall A, B \in SCOI(U)$. For all $x \in U$,
$$(A \cup B)(x) = max(A(x), B(x)),$$
$$(A \cap B)(x) = min(A(x), B(x)).$$

Based on the above definitions, we can prove that SCOI has the following operations and properties.

**Property 1.** $\forall A, B, C \in SCOI(U)$.

(1) Idempotent law: $A \cup A = A$, $A \cap A = A$

(2) Commutative law: $A \cup B = B \cup A$, $A \cap B = B \cap A$

(3) Associative law: $(A \cup B) \cup C = A \cup (B \cup C)$, $(A \cap B) \cap C = A \cap (B \cap C)$

(4) Absorption law: $A \cap (A \cup B) = A$, $A \cup (A \cap B) = A$

(5) Distributive law: $A \cup (B \cap C) = (A \cup B) \cap (A \cup C)$, $A \cap (B \cup C) = (A \cap B) \cup (A \cap C)$

(6) 0-1 law: $A \cup \varnothing = A$, $A \cap \varnothing = \varnothing$, $U \cup A = U$, $U \cap A = A$

Proof: Same as the proof for classical sets.

Due to the fact that there is only one negation (complement) in both classical and fuzzy set theories, and SCOI distinguishes negation into contradictory negation, opposing negation, and intermediary negation, the operation on negation in SCOI has many more in-depth and interesting properties than in classical and fuzzy sets.

**Property 2.** $\forall A, B, C \in SCOI(U)$.

(1) $A \subseteq B$, if and only if $B^\neg \subseteq A^\neg$

(2) $A \subseteq B$, if and only if $B^\sim \subseteq A^\sim$

(3) $A^\sim \subseteq B^\sim$, if and only if $B^{\sim\sim} \subseteq A^{\sim\sim}$

(4) $A \subseteq B^\neg$, if and only if $B \subseteq A^\neg$

(5) $A^\neg \subseteq B$, if and only if $B^\neg \subseteq A$

Proof:

(1). Suppose $A \subseteq B$. $A(x) \leq B(x)$ according to the definition 2, i.e. $1 - B(x) \leq 1 - A(x)$. According to the definition 1, $B^\neg(x) \leq A^\neg(x)$. $B^\neg \subseteq A^\neg$ by the definition 2. Conversely, the same can be proved.

(2). Suppose $A \subseteq B$. $A(x) \leq B(x)$ according to the definition 2, $f^\sim(A(x)) \geq f^\sim(B(x))$ by the proposition 3, $A^\sim(x) \geq B^\sim(x)$ by the definition 1, $B^\sim \subseteq A^\sim$ by the definition 2. Conversely, the same can be proved.

(3). Suppose $A^\sim \subseteq B^\sim$. $A^\sim(x) \leq B^\sim(x)$ according to the definition 2. According to the proposition 3, $f^\sim(A^\sim(x)) \geq f^\sim(B^\sim(x))$. $A^{\sim\sim}(x) \geq B^{\sim\sim}(x)$ by the definition 1, $B^{\sim\sim} \subseteq A^{\sim\sim}$ by the definition 2. Conversely, the same can be proved.

(4). Suppose $A \subseteq B^\neg$. By the definition 2, $A(x) \leq B^\neg(x)$. According to the definition 1, $A(x) \leq 1 - B(x)$, i.e. $B(x) \leq 1 - A(x) = A^\neg(x)$. $B \subseteq A^\neg$ by the definition 2. Conversely, the same can be proved.

(5). the proof as (4). □

**Property 3.** $\forall A, B, C \in SCOI(U)$.

(1) $A^{\neg\neg} = A$

(2) $A^\sim = A^{\neg\sim}$

(3) $A^\neg = A^\neg \cup A^\sim$

(4) $A^\sim = A^\neg \cap A^{\neg\neg}$

(5) $A^{\neg\neg} = A \cup A^\sim$

(6) $(A \cup B)^\neg = A^\neg \cap B^\neg$

(7) $(A \cap B)^\neg = A^\neg \cup B^\neg$

Proof:

(1). According to the definition 1, $A^{\neg\neg}(x) = 1 - A^\neg(x) = 1 - (1 - A(x)) = A(x)$. $A^{\neg\neg} = A$ by the definition 2.

(2). If $A^\sim(x) \geq A^{\neg\sim}(x)$, then $(A^\neg)^\sim(x) \geq (A^\neg)^{\neg\sim}(x) = A^{\neg\neg\sim}(x)$. Because of $A^{\neg\neg} = A$, so $A^{\neg\sim}(x) \geq A^\sim(x)$. Conversely, if $A^\sim(x) \leq A^{\neg\sim}(x)$ then $(A^\neg)^\sim(x) \leq (A^\neg)^{\neg\sim}(x) = A^{\neg\neg\sim}(x)$, i.e. $A^{\neg\sim}(x) \leq A^\sim(x)$. Therefore, $A^\sim(x) = A^{\neg\sim}(x)$. $A^\sim = A^{\neg\sim}$ by the definition 2.

(3). According to the definition 1 and the definition 3, $A^\neg(x) = max(A^\neg(x), A^\sim(x)) = (A^\neg \cup A^\sim)(x)$. By the definition 2, $A^\neg = A^\neg \cup A^\sim$.

(4). According to the definition 1 and the definition 3, $(A^\neg \cap A^{\neg\neg})(x) = min(A^\neg(x), A^{\neg\neg}(x)) = min([max(A^\neg(x), A^\sim(x))], [max(A^{\neg\neg}(x), A^{\neg\sim}(x))])$. Due to $A^{\neg\neg} = A$, so $(A^\neg \cap A^{\neg\neg})(x) = min([max(A^\neg(x), A^\sim(x))], [max(A(x), A^\sim(x))])$. Among them, if $A(x) \geq A^\sim(x)$, then $A^\sim(x) \geq A^\neg(x)$ by the proposition 1. Thus, $(A^\neg \cap A^{\neg\neg})(x)$

= $A\tilde{}(x)$; if $A(x) \leq A\tilde{}(x)$, $A\tilde{}(x) \leq A^{\daleth}(x)$ by the proposition 1. Thus, $(A^{\neg}\cap A^{\daleth\neg})(x) = A\tilde{}(x)$. By the definition 2, $A\tilde{} = A^{\neg} \cap A^{\daleth\neg}$.

(5). According to (3), $A^{\daleth\neg} = A^{\daleth\daleth}\cup A^{\daleth\tilde{}}$. From (1) and (2), $A^{\daleth\neg} = A\cup A\tilde{}$.

(6). According to the definition 1 and the definition 3, $(A\cup B)^{\daleth}(x) = 1 - (A\cup B)(x) = 1 - max(A(x), B(x))$, $(A^{\daleth}\cap B^{\daleth})(x) = min(A^{\daleth}(x), B^{\daleth}(x)) = min(1-A(x), 1-B(x))$. Among them, if $A(x) \geq B(x)$, then $(A\cup B)^{\daleth}(x) = 1-A(x)$ and $(A^{\daleth}\cap B^{\daleth})(x) = 1-A(x)$; if $A(x) \leq B(x)$, then $(A\cup B)^{\daleth}(x) = 1-B(x)$ and $(A^{\daleth}\cap B^{\daleth})(x) = 1-B(x)$. Thus, $(A\cup B)^{\daleth}(x) = (A^{\daleth}\cap B^{\daleth})(x)$. By the definition 2, $(A\cup B)^{\daleth} = A^{\daleth}\cap B^{\daleth}$.

(7). According to the definition 1 and the definition 3, $(A\cap B)^{\daleth}(x) = 1-(A\cap B)(x) = 1-min(A(x), B(x))$, $(A^{\daleth}\cup B^{\daleth})(x) = max(A^{\daleth}(x), B^{\daleth}(x)) = max(1-A(x), 1-B(x))$. Among them, if $A(x) \geq B(x)$, then $(A\cap B)^{\daleth}(x) = 1-B(x)$ and $(A^{\daleth}\cup B^{\daleth})(x) = 1-B(x)$; if $A(x) \leq B(x)$, then $(A\cap B)^{\daleth}(x) = 1-A(x)$ and $(A^{\daleth}\cup B^{\daleth})(x) = 1-A(x)$. Thus, $(A\cap B)^{\daleth}(x) = (A^{\daleth}\cup B^{\daleth})(x)$. By the definition 2, $(A\cap B)^{\daleth} = A^{\daleth}\cup B^{\daleth}$. □

In SCOI, if a set is the fuzzy set, it can be proven that SCOI has the following properties.

**Property 4**. $\forall A \in SCOI(U)$. If $A$ is a fuzzy set, $\Delta, \nabla \in \{\daleth, \sim, \neg\}$, then

(1) $A\cup A^{\Delta} \neq U$

(2) $A\cup A^{\nabla} \neq U$

(3) $A^{\Delta}\cup A^{\nabla} \neq U$

(4) $A\cap A^{\Delta} \neq \emptyset$

(5) $A\cap A^{\nabla} \neq \emptyset$

(6) $A^{\Delta}\cap A^{\nabla} \neq \emptyset$

For the fuzzy sets in SCOI, the property 3 and property 4 shows that Law of Excluded Middle and Law of Contradiction are not valid.

### 4.1.2 Three kinds of negation in SCOI are different fuzzy negations

In the classical set theory, the range of the characteristic function of a set is {0, 1}. For the fuzzy sets (Zadeh, 1965), it essentially expands the range of characteristic functions of classical sets from {0, 1} to [0, 1]. A fuzzy set is determined by a mapping (membership function) from the domain $X$ to [0, 1]. Therefore, the negation of a fuzzy set (called "fuzzy negation") is a mapping from [0, 1] to [0, 1].

In the fuzzy set theory, the definition of fuzzy negation is as follows (Wang, 1999; Trillas, 1979):

**Definition 4**. A function $N$: [0, 1] $\rightarrow$ [0, 1] is a fuzzy negation, if

(N1) $N(1) = 0$ and $N(0) = 1$;

(N2) if $x \leq y$ then $N(y) \leq N(x)$, $\forall x, y \in [0, 1]$;

(1) $N$ is a *strict* fuzzy negation, if

(N3) $N$ is continuous;

(N4) if $x < y$ then $N(y) < N(x)$, $\forall x, y \in [0, 1]$;

(2) $N$ is a *strong* fuzzy negation, if

(N5) $N(N(x)) = x$, $\forall x \in [0, 1]$;

(3) If there is $e \in [0, 1]$ such that $N(e) = e$, then $e$ is called the equilibrium point of $N$.

For SCOI, according to the definition of SCOI, if $A$ is a fuzzy set, the contradictory negation $\neg$, opposite negation $\daleth$ and intermediary negation $\sim$ of $A$ in SCOI are functions $f^{\neg}$, $f^{\daleth}$ and $f^{\tilde{}}$ from [0, 1] to [0, 1], respectively. According to Definition 4, it can be proven that $\daleth$, $\sim$ and $\neg$ are three fuzzy negations with different characteristics.

**Proposition 4**. The opposite negation $\daleth$ in SCOI is a strict and strong fuzzy negation.

Proof: $\forall x, y \in [0, 1]$. According to Definition 1, $f^{\daleth}(x) = 1-x$. Because $f^{\daleth}(1) = 0$, $f^{\daleth}(0) = 1$, $f^{\daleth}$ satisfies (N1) in Definition 4. By the proposition 4, $f^{\daleth}$ is a decreasing function, that is, if $x \leq y$ then $f^{\daleth}(y) = 1-y \leq f^{\daleth}(x) = 1-x$, if $x < y$ then $f^{\daleth}(y) = 1-y < f^{\daleth}(x) = 1-x$. Moreover, as defined by SCOI, $f^{\daleth}$ is a continuous function, $f^{\daleth}(f^{\daleth}(x)) = 1-(1-x) = x$. So, $f^{\daleth}$ satisfies (N2), (N3), (N4) and (N5) in Definition 4. According to Definition 4, the opposite negation $\daleth$ in SCOI is a strict and strong fuzzy negation. □

**Proposition 5**. The intermediary negation $\sim$ in SCOI is a strict fuzzy negation.

Proof: $\forall x, y \in [0, 1]$. According to Definition 1, there are only two cases (a) and (d) for $f^\sim(x) = 1$ in Definition 1. If the case (a), then $f^\sim(1) = 1-\lambda$ when $x = 1$. Since $\lambda \in [½, 1)$ and $f^\sim$ is continuous function, the limit of $f^\sim(1)$ is 0 when $\lambda \to 1$. If the case (d), then $f^\sim(1) = \lambda$ when $x = 1$. Since $\lambda \in (0, ½]$ and $f^\sim$ is continuous function, the limit of $f^\sim(1)$ is 0 when $\lambda \to 0$. for the same reason, there are only two cases (b) and (c) for $f^\sim(x) = 0$ in Definition 1. If the case (b), then $f^\sim(0) = \lambda$ when $x = 0$. Since $\lambda \in [½, 1)$ and $f^\sim$ is continuous function, the limit of $f^\sim(0)$ is 1 when $\lambda \to 1$. If the case (c), then $f^\sim(0) = 1-\lambda$ when $x = 0$. Since $\lambda \in (0, ½]$ and $f^\sim$ is continuous function, the limit of $f^\sim(0)$ is 1 when $\lambda \to 0$. Therefore, $f^\sim$ satisfies (N1) in Definition 4. Moreover, by the proposition 3, $f^⇁$ is a decreasing function, that is, if $x \leq y$ then $f^\sim(y) \leq f^\sim(x)$, if $x < y$ then $f^\sim(y) < f^\sim(x)$. Therefore, $f^\sim$ satisfies (N2), (N3) and (N4) in Definition 4. According to Definition 4, the intermediary negation ~ in SCOI is a strict fuzzy negation. □

**Proposition 6.** The contradictory negation ¬ in SCOI is a strict fuzzy negation.

Proof: According to Definition 1, $f^\neg(x) = max(f^⇁(x), f^\sim(x))$. By the proposition 4 and proposition 5, $f^⇁$ is a strict and strong fuzzy negation, and $f^\sim$ is a strict fuzzy negation. So, $f^\neg$ has the common properties of $f^⇁$ and $f^\sim$. Therefore $f^\neg$ is a strict fuzzy negation. □

**Proposition 7.** The contradictory negation¬, opposite negation ⇁ and intermediary negation~ in SCOI have a same equilibrium point ½.

Proof: Because of $f^\neg(½) = f^⇁(½) = f^\sim(½) = ½$ in SCOI, by the definition 4, ½ is an equilibrium point for the contradictory negation¬, opposite negation ⇁ and intermediary negation~. □

### 4.2 LCOI: Logic with contradictory negation, opposite negation and intermediary negation

In this section, based on SCOI, we further propose a "logic with contradictory negation, oppositional negation and mediated negation". We still use the symbols ¬, ⇁, and ~ denote "contradictory negation", "opposing negation" and "intermediary negation" respectively. In addition, the symbols ∨, ∧ and → denote "disjunction", "conjunction" and "implication" respectively. The symbol "⊢" denote formal deduction.

**Definition 1.** Let ℑ be set of atomic proposition. $\forall A \in ℑ$, A is called well-formed formula (or formula). If A, B are formulas, then ¬A, ⇁A, ~A, A→B, A∨B and A∧B are formulas.

(I) The following formulas as axioms:
   (a1)  A→(B→A)
   (a2)  (A→(A→B))→(A→B)
   (a3)  (A→B)→((B→C)→(A→C))
   (a4)  (A→ ¬B)→(B→ ¬A)
   (a5)  (A→ ⇁B)→(B→ ⇁A)
   (a6)  ¬A→(A→B)
   (a7)  ((A → ¬A)→B)→((A→B)→ B)
   (a8)  A → A ∨ B
   (a9)  B → A ∨ B
   (a10) A ∧ B → A
   (a11) A ∧ B→B
   (a12) ⇁ A → ¬A ∧ ¬ ~A
   (a13) ~A → ¬A ∧ ¬ ⇁ A

(II) The deduction rules:
   [I1]  $A_1, A_2, …, A_n ⊢ A_i$ (1≤ $i$ ≤ n)
   [I2]  A→B, A ⊢ B

The propositional calculus formal system determined above is called "logic with contradictory negation, opposite negation and intermediary negation", for short LCOI.

According to the operational properties of SCOI ((3) in the property 3), the contradictory negation set $A^\neg$ of set $A$ in SCOI has the following relationship with the opposite negation $A^⇁$ and the intermediary negation $A^\sim$ of $A$:

$$A^{\neg} = A^{\daleth} \cup A^{\sim}$$

Correspondingly, in LCOI, we can define the relationship between the contradictory negation ¬A and the opposite negation ⅂A and the intermediary negation ~A of the formula A as follows:

**Definition 2**. In LCOI,

$$\neg A = \daleth A \vee \sim A.$$

*Remark* 1. It can be proved that (i) some axioms and connectives in LCOI are not independent. (ii) the connectives set {¬, ⅂, ~, ∨, ∧, →} can be reduced to {⅂, ~, ∨, →}. Moreover, {¬, ⅂, ~, ∨, ∧, →} can be reduced to {⅂, ~, →} if ∨ and ∧ are defined by A∨B = ⅂A → B and A∧B = ⅂(A → ⅂B). That is ⅂, ~ and → are original connectives in LCOI.

*Remark* 2. In LCOI, A → B is not equivalent to ¬A∨B and ⅂A∨B.

### 4.2.1 Formal inference relations and meaning in LCOI

In LCOI, since "negation" is divided into "contradictory negation", "opposite negation" and "intermediary negation", in addition to maintaining the basic inference relations of the general logic, LCOI also has many special formal inference relations (formal theorems). It can be considered that these formal inference relations embody LCOI's ability to reflect the laws of knowledge reasoning. We list some main results as follows.

**Theorem 1**. Let A, B and C be the formulas in LCOI. Then

[1]    ⊢ A→A  
[2]    ⊢ ((A→B)→C)→(B→C)  
[3]    ⊢ A→((A→B)→(C→B))  
[4]    ⊢ A→((A→B)→B)  
[5]    ⊢ (((A→B)→B)→C)→(A→C)  
[6]    ⊢ (A→(B→C))→(B→(A→C))  
[7]*    ⊢ (B→C)→((A→B)→(A→C))  
[8]    ⊢ ((A→B)→(A→(A→C)))→((A→B)→(A→C))  
[9]    ⊢ (B→(A→C))→((A→B)→(A→C))  
[10] ⊢ (A→(B→C))→((A→B)→(A→C))

Proof [1]:  
(1) (A→(A→A))→(A→A)      (a2)  
(2) A→(A→A)      (a1)  
(3) A→A      (1)(2)[I2]  

Proof [2]:  
(1) (B→(A→B))→(((A→B)→C)→(B→C))      (a3)  
(2) B→(A→B)      (a1)  
(3) ((A→B)→C)→(B→C)      (1)(2)[I2]  

Proof [3]:  
(1) (C→A)→((A→B)→(C→B))→(A→((A→B)→(C→B)))      [2]  
(2) (C→A)→((A→B)→(C→B)      (a3)  
(3) A→((A→B)→(C→B))      (1)(2)[I2]  

Proof [4]:  
(1) (A→((A→B)→((A→B)→B)))→  
((((A→B)→((A→B)→B))→((A→B)→B))→(A→((A→B)→B)))      (a3)  
(2) A→((A→B)→((A→B)→B))      [3]  
(3) ((A→B)→((A→B)→B))→((A→B)→B))→(A→((A→B)→B))      (1)(2)[I2]  
(4) (A→B)→(((A→B)→B)→((A→B)→B))      [3]  
(5) A→((A→B)→B)      (3)(4)[I2]  

The same method can prove [5]−[10].

**Theorem 2**. Let A, B be the formulas in LCOI. Then

[11]      ⊢ ¬A→(A→ ¬(B→B))

- [12] ⊢ B→((A→¬B)→¬A)
- [13] ⊢ (A→¬B)→((A→B)→¬A)
- [14]* ⊢ (A→B)→((A→¬B)→¬A)   (law of reduction to absurdity)
- [15]* ⊢ (A→B)→(¬B→¬A)
- [16] ⊢ (A→¬A)→¬A
- [17]* ⊢ A→¬¬A
- [18]* ⊢ (¬A→¬B)→(B→A)
- [19] ⊢ ¬¬A→(¬¬A→A)
- [20]* ⊢ ¬¬A→A
- [21]* ⊢ (¬A→B)→((¬A→¬B)→A)   (law of indirect proof)

We choose to prove [11], [14]* and [17]*, and the rest can be proved by the same method.

Proof [11]:
- (1) $((B{\to}B){\to}\neg A){\to}(A{\to}\neg(B{\to}B))$  (a4)
- (2) $((B{\to}B){\to}\neg A){\to}(A{\to}\neg(B{\to}B)){\to}(\neg A{\to}(A{\to}\neg(B{\to}B)))$  [2]
- (3) $\neg A{\to}(A{\to}\neg(B{\to}B))$  (1)(2)[I2]

Proof [14]*:
- (1) $(A{\to}\neg B){\to}((A{\to}B){\to}\neg A)$  [13]
- (2) $((A{\to}\neg B){\to}((A{\to}B){\to}\neg A)){\to}((A{\to}B){\to}((A{\to}\neg B){\to}\neg A))$  [6]
- (3) $(A{\to}B){\to}((A{\to}\neg B){\to}\neg A)$  (1)(2)[I2]

Proof [17]*:
- (1) ¬A→¬A   [1]
- (2) (¬A→¬A)→(A→¬¬A)   (A4)
- (3) A→¬¬A   (1)(2)[I2]

The above theorem 1 and theorem 2 show that L<sub>COI</sub> retains the basic properties of the general logic.

**Theorem 3.** Let A, B be the formulas in L<sub>COI</sub>. Then
- [1⁰] ⊢ ⊣A→(A→⊣(B→B))
- [2⁰] ⊢ B→((A→⊣B)→⊣A)
- [3⁰] ⊢ (A→⊣B)→(A→(B→⊣(B→B)))
- [4⁰] ⊢ (A→B)→((A→⊣B)→⊣A)   (new "law of reduction to absurdity")
- [5⁰] ⊢ (A→B)→(⊣B→⊣A)
- [6⁰] ⊢ (A→⊣A)→⊣A
- [7⁰] ⊢ A→⊣⊣A
- [8⁰] ⊢ (⊣A→⊣B)→(B→A)
- [9⁰] ⊢ ⊣⊣A→(⊣⊣A→A)
- [10⁰] ⊢ ⊣⊣A→A

Proof [1⁰]:
- (1) $((B{\to}B){\to}\mathord{\dashv} A){\to}(A{\to}\mathord{\dashv}(B{\to}B))$  (a5)
- (2) $((B{\to}B){\to}\mathord{\dashv} A){\to}(A{\to}\mathord{\dashv}(B{\to}B)){\to}(\mathord{\dashv} A{\to}(A{\to}\mathord{\dashv}(B{\to}B)))$  [2]
- (3) $\mathord{\dashv} A{\to}(A{\to}\mathord{\dashv}(B{\to}B))$  (1)(2)[I2]

The proof of [2⁰]–[10⁰] is like the proof of [1⁰]. In the proof of [2⁰]–[10⁰], replacing (a4) with (a5) can obtain the proof.

The theorem 3 presents the distinctive inference properties of L<sub>COI</sub>. Among them, the new "law of reduction to absurdity" ([4⁰]) indicates that L<sub>COI</sub> has expanded the meaning of the law of reduction to absurdity ([[14]*).

**Theorem 4.** Let A be the formula in L<sub>COI</sub>. Then
- [11⁰] ⊢ ¬(A∧¬A)   (Law of non-contradiction)
- [12⁰] ⊢ ¬(⊣A∧~A)   (Law of non-contradiction 1)
- [13⁰] ⊢ ¬(A∧~A)   (Law of non-contradiction 2)
- [14⁰] ⊢ ¬(A∧⊣A)   (Law of non-contradiction 3)

Proof [11⁰]:

(1)  A∧¬A→A                (a10)
    (2)  A∧¬A→¬A               (a11)
    (3)  ¬(A∧¬A)               [14](1)(2)[I2]
  Proof [12⁰]:
    (1)  ⊣A∧~A→⊣A              (a10)
    (2)  ⊣A∧~A→~A              (a11)
    (3)  ⊣A→¬A∧¬~A             (a12)
    (4)  ⊣A∧~A→¬A∧¬~A          (1)(3)(a3)[I2]
    (5)  ¬A∧¬~A→¬~A            (a11)
    (6)  ⊣A∧~A→¬~A             (4)(5)(a3)[I2]
    (7)  ¬(⊣A∧~A)              (2)(6)[14][I2]
  Proof [13⁰]:
    (1)  A∧~A→A                (a10)
    (2)  A∧~A→~A               (a11)
    (3)  ~A→¬A∧¬⊣A             (a13)
    (4)  A∧~A→¬A∧¬⊣A           (2)(3)(a3)[I2]
    (5)  ¬A∧¬⊣A→¬A             (a10)
    (6)  A∧~A→¬A               (4)(5)(a3)[I2]
    (7)  ¬(A∧~A)               (1)(6)[14][I2]
  Proof [14⁰]:
    (1)  A∧⊣A→A                (a10)
    (2)  A∧⊣A→⊣A               (a11)
    (3)  ⊣A→¬A∧¬~A             (a12)
    (4)  A∧⊣A→¬A∧¬~A           (2)(3)(a3)[I2]
    (5)  ¬A∧¬⊣A→¬A             (a10)
    (6)  A∧⊣A→¬A               (4)(5)(a3)[I2]
    (7)  ¬(A∧⊣A)               (1)(6)[14][I2]

The theorem 4 shows that L_COI maintains the "law of non-contradiction" in the general logic ([11⁰]) and adds three new laws of non-contradiction ([12⁰], [13⁰], [14⁰]). In other words, L_COI extends the meaning of "contradiction" in general logic. Not only A and ¬A is contradictory in L_COI, but also any two in {A, ⊣A, ~A}.

We can also prove the following formal inference relation peculiar to L_COI.

**Theorem 5**.  Let A, B be the formulas in L_COI. Then
    (1)  A ⊢ A
    (2)  A ⊢ ⊣⊣A
    (3)  ⊣⊣A ⊢ A
    (4)  A ⊢ ¬¬A
    (5)  ¬¬A ⊢ A
    (6)  ⊣A ⊢ A→B
    (7)  B ⊢ A→B
    (8)  ⊣(A→B) ⊢ A, ⊣B
    (9)  A, ⊣B ⊢ ⊣(A→B)
    (10) ~A ⊢ ~⊣A
    (11) ~⊣A ⊢ ~A
    (12) A ⊢ ¬⊣A ∧ ¬~A

We choose to prove (1)–(3) and (12), and the remainder can be proved by the same method.
  Proof (1):
  (a)  A
  (b)  A→(A→A)     (a1)
  (c)  A            (a)(b)[I2]
  Proof (2):

*(a)* $A$
*(b)* $A \to \neg\neg A$     $[7^0]$
*(c)* $\neg\neg A$     *(a)(b)[I2]*

Proof (3):
*(a)* $\neg\neg A$
*(b)* $\neg\neg A \to A$     $[10^0]$
*(c)* $A$     *(a)(b)[I2]*

Proof (12):
*(a)* $A$
*(b)* $\neg\neg A$     *(2)(3)*
*(c)* $\neg\neg A \to \neg\neg A \wedge \neg \sim \neg A$     *(a12)*
*(d)* $\neg\neg A \wedge \neg \sim \neg A$     *(b)(c)[I2]*
*(e)* $\neg\neg A \wedge \neg \sim A$     *(8)(9)*

**Theorem 6.** Let A, B be the formulas in L<small>COI</small>. Then

(13)    $A, \neg A \vdash B$
(14)    $A, \neg A \vdash B$
(15)    $A, \sim A \vdash B$
(16)    $\neg A, \sim A \vdash B$
(17)    $A \vdash \neg\neg \sim A$

Proof (13):
*(a)* $A$
*(b)* $\neg A$
*(c)* $\neg A \to (A \to B)$     *(a6)*
*(d)* $A \to B$     *(b)(c)[I2]*
*(e)* $B$     *(a)(d)[I2]*

Proof (14):
*(a)* $A$
*(b)* $\neg A$
*(c)* $\neg A \to \neg A \wedge \neg \sim A$     *(a12)*
*(d)* $\neg A \wedge \neg \sim A$     *(b)(c)[I2]*
*(e)* $\neg A \wedge \neg \sim A \to \neg A$     *(a10)*
*(f)* $\neg A$     *(d)(e)[I2]*
*(g)* $B$     *(a)(f)(12)*

Proof (15):
*(a)* $A$
*(b)* $\sim A$
*(c)* $\sim A \to \neg A \wedge \neg \neg A$     *(a13)*
*(d)* $\neg A \wedge \neg \neg A$     *(b)(c)[I2]*
*(e)* $\neg A \wedge \neg \neg A \to \neg A$     *(a10)*
*(f)* $\neg A$     *(d)(e)[I2]*
*(g)* $B$     *(a)(f)(12)*

Proof (16):
*(a)* $\neg A$
*(b)* $\sim A$
*(c)* $\neg A \to \neg A \wedge \neg \sim A$     *(a12)*
*(d)* $\neg A \wedge \neg \sim A$     *(a)(c)[I2]*
*(e)* $\neg A \wedge \neg \sim A \to \neg \sim A$     *(a11)*
*(f)* $\neg \sim A$     *(d)(e)[I2]*
*(g)* $B$     *(a)(f)(12)*

Proof (17):
*(a)* $A$
*(b)* $\neg\neg A \wedge \neg \sim A$     *(12)*
*(c)* $\neg\neg \sim A \wedge \neg \sim \sim A$     *(b)*
*(d)* $\neg\neg \sim A \wedge \neg \sim \sim A \to \neg\neg \sim A$     *(a10)*

    *(e)*   ¬⊣ ~A                           *(c)(d)*[I2]

Because of the arbitrariness of B, the theorem 6 shows that under the meaning of "contradiction" in LCOI, not only any conclusion can be deduced from the premises A and ¬A, but also any conclusion can be deduced from any two in {A, ⊣ A, ~A}.

### 4.2.2 Semantics of LCOI

Regarding the semantics of LCOI, we provide a three-valued semantic interpretation of LCOI in this section. Under this semantic interpretation, it is proved that LCOI is reliable and complete.

Let $\Sigma$ be a set of formulas in LCOI. Since LCOI is a formal logic, defining the formal deduction $\Sigma \vdash A$ is provable in LCOI, just as it is in other formal logic.

**Definition 1**. The formal deduction $\Sigma \vdash A$ ($\Sigma$ can be empty set) is provable in LCOI, if there exists a finite sequence of formulas $E_1, E_2, \ldots, E_n$ such that $E_n = A$ and for each $E_n$ ($1 \leq k \leq n$), either $E_k$ is an axiom in LCOI or $E_k$ follows from $E_i$ and $E_j$ ($i < k, j < k$) using the deduction rule in LCOI, then $E_1, E_2, \ldots, E_n$ is called a "*proof*" of $\Sigma \vdash A$, n *length* of *proof*. $\Sigma \vdash A$ is denoted $\vdash A$ when $\Sigma$ is empty.

**Definition 2** (*three-valued interpretation*). Let A, B be formulas in LCOI. Mapping $\partial : \Sigma \to \{0, ½, 1\}$ is called a three-valued assignment, if

(1) $\partial(A) + \partial(\dashv A) = 1$.

(2) $\partial(\sim A) = \begin{cases} ½, & \text{when } \partial(A) = 1 \\ 1, & \text{when } \partial(A) = ½ \\ ½, & \text{when } \partial(A) = 0 \end{cases}$

(3) $\partial(A \to B)$ is a binary function $\Re(\partial(A), \partial(B))$:

| $\Re$ | | $\partial(B)$ | | |
|---|---|---|---|---|
| | | 1 | ½ | 0 |
| $\partial(A)$ | 1 | 1 | ½ | 0 |
| | ½ | 1 | 1 | ½ |
| | 0 | 1 | 1 | 1 |

(4) $\partial(A \land B) = \min(\partial(A), \partial(B))$.

(5) $\partial(\neg A) = \max(\partial(\dashv A), \partial(\sim A))$.

**Definition 3** (*three-valued valid formula*). Let $\Sigma$ be a set of formulas in LCOI, and A be a formula in LCOI. (1) If $\partial(A) = 1$ for any three-valued assignment $\partial$ then A is called the three-valued valid formula, which is denoted $\vDash A$. (2) For any $B \in \Sigma$, if there is three-valued assignment $\partial$ such that $\partial(B) = 1$, then $\Sigma$ is called three-valued satisfiable. (3) If any three-valued assignment that satisfies $\Sigma$ certain satisfy A, then $\Sigma \vdash A$ is called the three-valued valid deduction, denoted as $\Sigma \vDash A$.

**Theorem 1** (*Soundness theorem*). Let $\Sigma$ be a set of formulas in LCOI, and A be a formula in LCOI.

(a) If $\vdash A$, then $\vDash A$.

(b) If $\Sigma \vdash A$, then $\Sigma \vDash A$.

Proof: Suppose $\vdash A$. By Definition 1, there exists a proof of $\vdash A$: $E_1, E_2, \ldots, E_n$, where $E_n = A$. We will prove by induction on the length n of $E_1, E_2, \ldots, E_n$.

(i) When $n = 1$, $E_1 = A$ by the definition. Namely, A is an axiom in LCOI. Since the axioms in LCOI are all valid formulas, $\vdash A$ is a three-valued valid deduction. So $\vDash A$ by the definition 3.

(ii) Suppose (a) holds when $n < k$.

When $n = k$, according to the definition 1, either $E_n$ is an axiom in LCOI or $E_n$ follows from $E_i$ and $E_j$ ($i < n, j < n$) using the deduction rule in LCOI. If $E_n$ is an axiom in $E_n$, (a) holds by (i); If $E_n$ follows from $E_i$ and $E_j$ ($i < n, j < n$) using the deductions rule, then $E_i$ and $E_j$ are three-valued valid formulas according to the induction hypothesis, so $E_n$ is three-valued valid formula. Namely $\vdash A$ is three-valued valid deduction. So $\vDash A$ by the definition 3.

Therefore, by (i) and (ii), (a) is proved. (b) can be proved in the same way. □

**Definition 4** (*Maximal consistent set*). Let $\Sigma$ be a set of formulas in LCOI. If $\Sigma \vdash A$ is provable in LCOI for any formula A, then $\Sigma$ is called inconsistent set. If $\Sigma$ is a consistent set and has $\Sigma' = \Sigma$ for any consistent set $\Sigma' \supseteq \Sigma$, then $\Sigma$ is called *maximal consistent set*.

**Lemma 1**. For any consistent set $\Sigma$, there exists a maximal consistent set $\Sigma^*$ and $\Sigma \subseteq \Sigma^*$.

Proof: Let $A_1, A_2, \ldots, A_n, \ldots$ be all the formulas in LCOI. Defining $\Sigma_n$: $\Sigma_1 = \Sigma$, if $\Sigma_n \cup \{A_n\}$ is consistent then $\Sigma_{n+1} = \Sigma_n \cup \{A_n\}$, or else $\Sigma_{n+1} = \Sigma_n$. Let $\Sigma^* = \bigcup_{n=1}^{\infty} \Sigma_n$. $\Sigma^*$ can be verified as a maximal consistent set that $\Sigma^* \supseteq \Sigma$ by the definition 4. □

**Lemma 2**. Let $\Sigma$ be a set of formulas in LCOI. For the formula A in LCOI, if $A_1, A_2 \in \{A, \daleth A, \sim A\}$ and $A_1 \neq A_2$, then the following propositions are equivalent.

(1) $\Sigma$ is inconsistent set.
(2) $\Sigma \vdash A_1, A_2$ is provable in LCOI.
(3) $\Sigma \vdash A \wedge \neg A$ is provable in LCOI.

Proof: If $\Sigma$ is an inconsistent set, by the definition 4, $\Sigma \vdash A$ is provable in LCOI for any formula A. Since $A_1, A_1 \in \{A, \daleth A, \sim A\}$ that is $A_1$ and $A_1$ are formulas, thus $\Sigma \vdash A_1, A_2$ is provable in LCOI. Therefore, (2) can be deduced from (1). If $\Sigma \vdash A_1, A_2$ is provable in LCOI, since $A_1, A_1 \in \{A, \daleth A, \sim A\}$, thus $A_1, A_2 \vdash B$ is provable in LCOI according to (14)-(16) in theorem 6. Because of the arbitrariness of B, we can make B = $A \wedge \neg A$. Therefore, (3) can be deduced from (2). If $\Sigma \vdash A \wedge \neg A$ is provable in LCOI, according to [11⁰] in theorem: $\vdash \neg(A \wedge \neg A)$, so $\Sigma \vdash A \wedge \neg$ and $\vdash \neg(A \wedge \neg A)$. Since $A \wedge \neg A$ and $\neg(A \wedge \neg A)$ are contradictory, $\Sigma$ is an inconsistent set. Therefore, (1) can be deduced from (3). □

**Lemma 3**. Let $\Sigma$ be maximal consistent set. Then

(1) $\Sigma \vdash A$ is provable in LCOI, if and only if $A \in \Sigma$.
(2) If $A \vdash B$ and $B \vdash A$ are provable in LCOI, then $A \in \Sigma$ if and only if $B \in \Sigma$.

Proof: (1). Suppose $A \notin \Sigma$. Thus there are two kinds of case: either $\daleth A \in \Sigma$, or $\sim A \in \Sigma$. If $\daleth A \in \Sigma$, according to the deduction rules (I1), $\Sigma \vdash \daleth A$, that is $\Sigma \vdash \daleth A$ is provable in LCOI. Since $\Sigma$ is consistent set, thus $\Sigma \vdash A$ is unprovable in LCOI by the lemma 2. Similarly, if $\sim A \in \Sigma$, then $\Sigma \vdash \sim A$ according to the deduction rules (I1). That is $\Sigma \vdash \sim A$ is provable in LCOI. Since $\Sigma$ is consistent set, thus $\Sigma \vdash \sim A$ is unprovable in LCOI by the lemma 2. Conversely, if $A \in \Sigma$, then $\Sigma \vdash A$ according to the deduction rules (I1). That is $\Sigma \vdash A$ is provable in LCOI. (2). If $A \in \Sigma$, then $\Sigma \vdash A$ is provable in LCOI according to (1). Because of $A \vdash B$ is provable in LCOI, thus $\Sigma \vdash B$ is provable in LCOI, $B \in \Sigma$ according to (1). If $B \in \Sigma$, then $\Sigma \vdash B$ is provable in LCOI by (1). Because of $B \vdash A$ is provable in LCOI, thus $\Sigma \vdash A$ is provable in LCOI, $A \in \Sigma$ according to (1). □

**Lemma 4**. Let $\Sigma$ be maximal consistent set, A and B be any two formulas in LCOI. Then

(1) Only one of $A \in \Sigma$, $\daleth A \in \Sigma$ and $\sim A \in \Sigma$ holds.
(2) $A \to B \in \Sigma$, when $\daleth A \in \Sigma$ or $B \in \Sigma$.
(3) $\daleth (A \to B) \in \Sigma$, when $A \in \Sigma$ and $\daleth B \in \Sigma$.

Proof: (1). If there are two holds in $A \in \Sigma$, $\daleth A \in \Sigma$, $\sim A \in \Sigma$, then $\Sigma$ is an inconsistent set according to Lemma 2, this is in contradiction with the hypothesis. (2). If $\daleth A \in \Sigma$, then $\Sigma \vdash \daleth A$ by Lemma 3. According to (6) in Theorem 5: $\daleth A \vdash A \to B$, and then $\Sigma \vdash \daleth A \vdash A \to B$, that is $\Sigma \vdash A \to B$ is provable in LCOI. Thus, $A \to B \in \Sigma$ by Lemma 3. If $B \in \Sigma$, then $\Sigma \vdash B$ by Lemma 3. According to (7) in Theorem 5: $B \vdash A \to B$, thus $\Sigma \vdash B \vdash A \to B$. That is $\Sigma \vdash A \to B$ is provable in LCOI. Thus, $A \to B \in \Sigma$ by Lemma 3. (3). If $A \in \Sigma$ and $\daleth B \in \Sigma$, then $\Sigma \vdash A, \daleth B$ is provable in LCOI. According to (9) in Theorem 5: $A, \daleth B \vdash \daleth (A \to B)$, thus $\Sigma \vdash A, B \vdash \daleth (A \to B)$. That is $\Sigma \vdash \daleth (A \to B)$ is provable in LCOI. Therefore, $\daleth (A \to B) \in \Sigma$ by Lemma 3. □

**Definition 5**. Let $\Sigma$ be maximal consistent set. Based on the definition 2, if *p* is an atomic formula in LCOI, then

$$\partial(p) = \begin{cases} 1, & \text{when } p \in \Sigma. \\ \tfrac{1}{2}, & \text{when } \sim p \in \Sigma. \\ 0, & \text{when } \daleth p \in \Sigma. \end{cases}$$

**Lemma 5**. Let $\Sigma$ be maximal consistent set, and A be a formula in LCOI. Then there exists three-valued assignment $\partial$ such that

(1) $\partial(A) = 1$, if and only if $A \in \Sigma$.

(2) $\partial(A) = \tfrac{1}{2}$, if and only if $\sim A \in \Sigma$.

(3) $\partial(A) = 0$, if and only if $\daleth A \in \Sigma$.

Proof: Since formula A involves three conjunctions $\daleth$, $\sim$ and $\rightarrow$, we prove it inductively for the total number $n$ of occurrences of conjunctions in A.

When $n = 0$, i.e. A is an atomic formula in LCOI. Thus, (1), (2) and (3) holds according to Definition 5.

Suppose (1), (2) and (3) holds when $n < k$. When $n = k$, we prove as follows.

Let $A = \daleth B$. According to the definition 2, (i) $\partial(\daleth B) = 1$ if and only if $\partial(B) = 0$. $\daleth B \in \Sigma$ by Definition 5, that is $A \in \Sigma$. (ii) $\partial(\daleth B) = \tfrac{1}{2}$ if and only if $\partial(B) = \tfrac{1}{2}$, so $\sim B \in \Sigma$ by Definition 5. $\Sigma \vdash \sim B$ by Lemma 3, $\Sigma \vdash \sim \daleth B$ by (10) in the theorem 5, $\sim \daleth B \in \Sigma$ by Lemma 3, that is $\sim A \in \Sigma$. (iii) $\partial(\daleth B) = 0$ if and only if $\partial(B) = 1$, so $B \in \Sigma$ by Definition 5. $\Sigma \vdash B$ by Lemma 3. $\vdash B \rightarrow \daleth \daleth B$ by [7⁰] in Theorem 3. $\Sigma \vdash \daleth \daleth B$ according to the deduction rules [I2], $\daleth \daleth B \in \Sigma$ by Lemma 3, that is $\daleth A \in \Sigma$.

Let $A = \sim B$. Proof in three parts: (i) According to the definition 2, $\sim B \in \Sigma$ by Definition 5, that is $A \in \Sigma$. (ii) In order to prove that $\partial(\sim B) = \tfrac{1}{2}$ if and only if $\sim \sim B \in \Sigma$, we may prove that $\partial(\sim B) \neq \tfrac{1}{2}$ if and only if $\sim \sim B \notin \Sigma$ as follows. Suppose $\partial(\sim B) \neq \tfrac{1}{2}$. According to the definition 2, $\partial(B) = \tfrac{1}{2}$, and then $\sim B \in \Sigma$ by Definition 5, $\sim \sim B \notin \Sigma$ by (1) in the lemma 4. Conversely, suppose $\sim \sim B \notin \Sigma$. According to (1) in the lemma 4, $\sim B \in \Sigma$ or $\daleth \sim B \in \Sigma$. If $\sim B \in \Sigma$, according to the definition 5, $\partial(B) = \tfrac{1}{2}$, and then $\partial(\sim B) = 1$ by Definition 2, that is $\partial(\sim B) \neq \tfrac{1}{2}$. If $\daleth \sim B \in \Sigma$, according to (1) in the lemma 4, $\sim B \notin \Sigma$ and $\sim \sim B \notin \Sigma$. Since $\sim B \notin \Sigma$ contradicts $\sim B \in \Sigma$, $\daleth \sim B \in \Sigma$ does not exist. (iii) In order to prove that $\partial(\sim B) = 0$ if and only if $\daleth \sim B \in \Sigma$, we may prove that $\partial(\sim B) \neq 0$ if and only if $\daleth \sim B \notin \Sigma$ as follows. Suppose $\daleth \sim B \notin \Sigma$, According to (1) in the lemma 4, $\sim B \in \Sigma$ or $\sim \sim B \in \Sigma$. If $\sim B \in \Sigma$, and then $\partial(B) = \tfrac{1}{2}$ by Definition 5. $\partial(\sim B) = 1 (\neq 0)$ by Definition 2. If $\sim \sim B \in \Sigma$, then $\sim B \notin \Sigma$ by (1) in the lemma 4, so $B \in \Sigma$ or $\daleth B \in \Sigma$, $\partial(B) = 1$ or $\partial(B) = 0$ by the definition 5. According to the definition 2, $\partial(\sim B) = \tfrac{1}{2} (\neq 0)$. Conversely, suppose $\partial(\sim B) \neq 0$, according to the definition 2, $\partial(\sim B) = 1$ or $\partial(\sim B) = \tfrac{1}{2}$, that is $\sim B \in \Sigma$ or $\sim \sim B \in \Sigma$ by the definition 5. According to (1) in the lemma 4, if $\sim B \in \Sigma$ then $\daleth \sim B \notin \Sigma$; if $\sim \sim B \in \Sigma$ then $\daleth \sim B \notin \Sigma$.

Let $A = B \rightarrow C$. Proof in three parts: (i) According to the definition 2, $\partial(B \rightarrow C) = 1$ if and only if $\partial(B) = 0$ or $\partial(C) = 1$. $\partial(B) = 0$ or $\partial(C) = 1$ that is $\daleth B \in \Sigma$ or $C \in \Sigma$ by the definition 5, and then $B \rightarrow C \in \Sigma$ according to (2) in the lemma 4. (ii) According to the definition 2, $\partial(B \rightarrow C) = 0$ if and only if $\partial(B) = 1$ and $\partial(C) = 0$. If $\partial(B) = 1$ and $\partial(C) = 0$, then $B \in \Sigma$ and $\daleth C \in \Sigma$ by the definition 5, $\daleth (B \rightarrow C) \in \Sigma$ according to (3) in the lemma 4. (iii) In order to prove that $\partial(B \rightarrow C) = \tfrac{1}{2}$ if and only if $\sim(B \rightarrow C) \in \Sigma$, we may prove $\partial(B \rightarrow C) \neq \tfrac{1}{2}$ if and only if $\sim(B \rightarrow C) \notin \Sigma$ as follows. Suppose $\partial(B \rightarrow C) \neq \tfrac{1}{2}$. According to the definition 2, $\partial(B \rightarrow C) \neq \tfrac{1}{2}$ if and only if $\partial(B \rightarrow C) = 1$ or $\partial(B \rightarrow C) = 0$. If $\partial(B \rightarrow C) = 1$ then $B \rightarrow C \in \Sigma$ by (i), so $\sim(B \rightarrow C) \notin \Sigma$ by (1) in the lemma 4. If $\partial(B \rightarrow C) = 0$ then $\daleth(B \rightarrow C) \in \Sigma$ by (ii), so $\sim(B \rightarrow C) \notin \Sigma$ by (1) in the lemma 4. □

**Theorem 2** (*completeness theorem*). Let A and $\Sigma$ be a formula and a formula set in LCOI, respectively.

(a) If $\Sigma \vDash A$, then $\Sigma \vdash A$.

(b) If $\vDash A$, then $\vdash A$.

Proof: If $\Sigma \vdash A$ is unprovable in LCOI, $A \notin \Sigma$ according to the lemma 3, $\daleth A \in \Sigma$ or $\sim A \in \Sigma$ by the lemma 4. If $\daleth A \in \Sigma$, then there exists three-valued assignment $\partial$ such that $\partial(A) = 0$ by the lemma 5. If $\sim A \in \Sigma$, then there exists three-valued assignment $\partial$ such that $\partial(A) = \tfrac{1}{2}$ by the lemma 5, so $\partial(A) \neq 1$. According to the definition 3, $\Sigma \vDash A$ not holds. (b) is the situation of $\Sigma = \varnothing$ in (a), it can be proved in the same way. □

**Theorem 3** (*compactness theorem*). Let $\Sigma$ be an infinite set of formulas in LCOI. For any finite set $\Gamma \subset \Sigma$, if $\Gamma$ is consistent set then $\Sigma$ is consistent set.

Proof: Suppose $\Sigma$ is inconsistent set. $\Sigma \vdash A \wedge \neg A$ is provable in LCOI by the lemma 2. According to the definition 1, there exist finite set Γ = {$E_1$, $E_2$, …, $E_n$} and $E_n$ = A∧¬A in Lcoɪ. Γ ⊢ A∧¬A by the deduction rules [I1]. Since Σ is infinite set of formulas in Lcoɪ, thus Γ⊂Σ. According to [11⁰] in Theorem 4: ⊢ ¬(A ∧¬A), so Γ ⊢ A ∧¬A is contradictory with ⊢ ¬(A ∧¬A). Therefore, Γ is an inconsistent set. □

## 4.3 The objectivity and expressive power of "intermediary negation" in Scoɪ and Lcoɪ

In Scoɪ and Lcoɪ, in addition to contradiction negation and opposite negation, "intermediary negation" is the third form of negation that exists between two opposing concepts. Whether this form of negation exists in objective reality, and what kind of knowledge expression ability it has, we will demonstrate through the following examples.

In practice, the following types of descriptive statements are commonly found:
(1) *Clear statements*: "*x* is neither a positive integer nor a negative integer", "this profession is neither safe nor dangerous", "this country belongs to neither the East nor the West", and so on.
(2) *Fuzzy statements*: "*S* is neither happy nor unhappy", "my father is neither fat nor thin", "there is neither more nor less salt in the dish", and so on.

Obviously, this type of statement contains more complex negativity.

In Scoɪ, for the contradictory negation $A^¬$, opposite negation $A^~$ and intermediary negation $A^⇁$ of the set *A*, we has proved the following property (Property 3 in Section 4.1.1):

$$A^~ = A^¬ \cap A^{⇁¬} \qquad (*)$$

That is, $A^~$ is intersection of $A^¬$ and the contradictory negation $A^{⇁¬}$ of $A^⇁$.

To understand $A^~ = A^¬ \cap A^{⇁¬}$, Let's take the statement "*x* is neither a positive integer nor a negative integer" from (1) as an example, which is denoted as *Ex*.

According to Scoɪ, in *Ex*, non-positive integer, negative integer, and zero are respectively the contradictory negation, opposite negation, and intermediary negation of "positive integer", while "non-negative integer" is the contradictory negation of negative integers. They can be formally expressed as follows:

*A*: positive integer; $A^¬$: non positive integer; $A^⇁$: negative integer; $A^{⇁¬}$: non negative integer; $A^~$: zero

Based on Fig.1-Fig. 3 in Section 3.3, the extensional relationship between *A*, $A^¬$, $A^⇁$, $A^~$ and $A^{⇁¬}$ can be illustrated as follows:

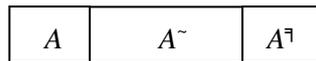

**Fig. 8.** The extensional relationships of *A*, $A^⇁$, $A^~$

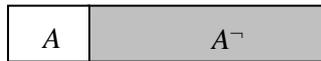 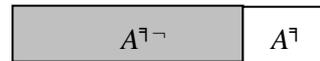

**Fig. 9**. The extensional relationship of *A* and $A^¬$    **Fig. 10**. The extensional relationship of *A* and $A^{⇁¬}$

According to the meaning of the statement *Ex*, *Ex* should be the common part of $A^¬$ in Figure 10 and $A^{⇁¬}$ in Figure 11, namely the intersection of $A^¬$ and $A^{⇁¬}$. That is, $A^~$ in Figure 9. It can be illustrated as follows:

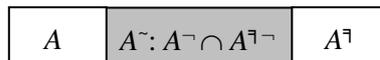

**Fig. 11**. $A^~$ is the intersection of $A^¬$ and $A^{⇁¬}$

According to the relationship between set theory and mathematical logic, a logical statement corresponds to a set. The property (∗) in Scoɪ can be expressed in Lcoɪ as follows:

$$\sim A = \neg A \wedge \neg ⇁ A \qquad (**)$$

where, ~A, ¬A, and ¬ ⊣ A are formulas in LCOI , A is atom formula.

For *Ex*, we can express it formally as follows:

A(*x*): *x* is positive integer, (i.e., A(*x*) is atom formula)

¬A(*x*): *x* is not positive integer, (i.e., contradictory negation of A(*x*))

⊣ A(*x*): *x* is negative integer, (i.e., opposite negation of A(*x*))

¬ ⊣ A(*x*): *x* is not negative integer, (i.e., contradictory negation of ⊣ A(*x*))

~A(*x*): *x* is zero, (i.e., intermediary negation of A(*x*))

Based on the meaning of *Ex*, its formal representation is as follows:

$$Ex: \neg A(x) \land \neg \dashv A(x)$$

That is, *Ex* is conjunctions of ¬A(*x*) and ¬ ⊣ A(*x*).

According to (∗∗), *Ex* = ~A(*x*), which means *Ex* is the intermediary negation of A(*x*). In other words, the statement *Ex* has the same meaning as "*x* is zero".

Regarding the statement *Ex*, as well as the different negations within *Ex* and their relationships, we can illustrate it as follows (Figure 12):

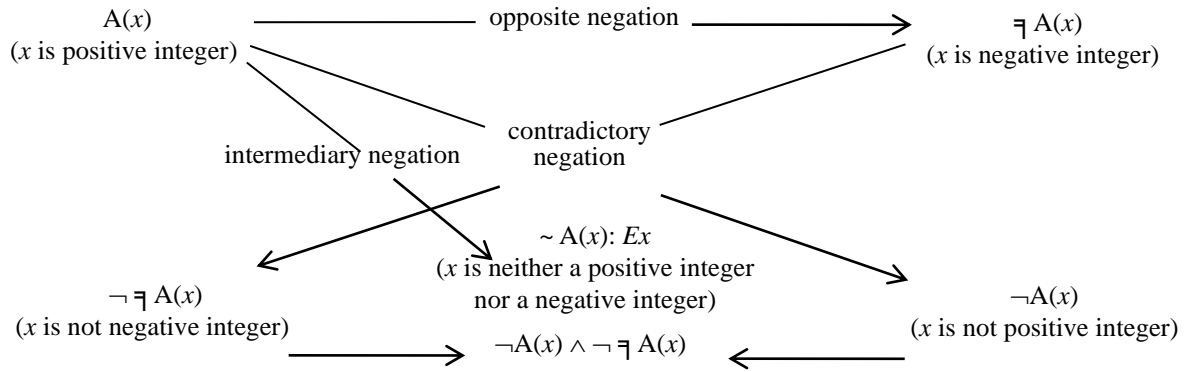

**Fig. 12**. The statement *Ex*, along with the different negations within *Ex* and their relationships

The above example uses the statement "*x* is neither a positive integer nor a negative integer" to demonstrate that this statement can be expressed through 'intermediary negation' in both SCOI and LCOI. Therefore, it indicates that:

- For statements of types (1) and (2) with more complex negativity, whether clear sentence and fuzzy sentence, they can be expressed by the "intermediary negation" of a set in SCOI and the "intermediary negation" of an atom formula in LCOI.
- The "intermediary negation" in SCOI and LCOI, it exists in objective reality.

In knowledge representation and natural language processing, for statements of types (1) and (2), if they are formalized using other methods, the expression will be more complex due to the inability to distinguish the three different negations and their relationships that exist in the knowledge.

## 5. Application

In this section, we apply SCOI to solve a multi-attribute decision-making problem to verify the effectiveness of SCOI and LCOI in practical applications.

**Example**: In China, an individual (person or household) decides whether to save the remaining money in the bank or spend it after meeting their daily living expenses. This decision depends on how much economic income the individual has each month and how much savings they have in the bank.

Suppose there are the following decision options (alternative schemes):

(a) If an individual has low savings, the remaining income after living expenses will be deposited to the bank each month.

(b) If an individual has high savings and high monthly income, the remaining income after living expenses is used for consumption each month.

(c) If an individual has high savings and moderate monthly income, a small amount of the remaining monthly income after living expenses will be deposited to the bank and the most portion is used for consumption each month.

(d) If an individual has moderate savings and moderate monthly income, most of the remaining monthly income after living expenses will be deposited to the bank and the small portion is used for consumption each month.

(*Remark*: If the individual's monthly income is low, it is considered that there is no remaining monthly income after monthly living expenses.)

**Question**: If individual M has a monthly income of 5,000 RMB and bank savings of 120,000 RMB, which of the four alternative schemes is the most suitable decision for M?

It is evident that this is an instance of fuzzy multi-attribute decision-making with different types of negativity. We will solve it using the following steps:

Step 1: Data integration; Step 2: Formal representation of the alternative schemes; Step 3: Solving the membership degree of data to fuzzy sets and their different negation sets; Step 4: Solve for the "threshold" of the range of values of the membership degree; Step 5: Inference; Step 6: Conclusions of decision-making.

(1) *Data integration*

In real life, the meaning of "high (or low) income" and "high (or low) deposit" is uncertain and people do not have unified understanding or awareness. One of the main influencing factors is the difference in people's living area. In 2007, we conducted a random survey of people in some areas of eastern China and obtained the following Table 1 (money unit: ￥/RMB).

**Table 1**. The understanding of "high (low) income" and "high (low) deposits" among people in eastern China

| views | City/Province | high income (monthly) | low income (monthly) | high deposit | low deposit |
|---|---|---|---|---|---|
| 1 | incity/Shanghai | ≥ 15,000 | ≤ 2,000 | ≥ 200,000 | ≤ 100,000 |
| 2 | Pudong/Shanghai | ≥ 20,000 | ≤ 2,500 | ≥ 250,000 | ≤ 150,000 |
| 3 | Xuhui/Shanghai | ≥ 10,000 | ≤ 2,000 | ≥ 200,000 | ≤ 80,000 |
| 2 | Nanjing/Jiangsu | ≥ 10,000 | ≤ 1,500 | ≥ 200,000 | ≤ 80,000 |
| 3 | Wuxi/Jiangsu | ≥ 12,000 | ≤ 1,200 | ≥ 150,000 | ≤ 100,000 |
| 4 | Suzhou/Jiangsu | ≥ 15,000 | ≤ 1,500 | ≥ 150,000 | ≤ 100,000 |
| 5 | Hefei/Anhui | ≥ 6,000 | ≤ 1,000 | ≥ 100,000 | ≤ 80,000 |
| 6 | Fuyang/Anhui | ≥ 5,000 | ≤ 1,000 | ≥ 100,000 | ≤ 50,000 |
| 7 | Tongning/Anhui | ≥ 4,000 | ≤ 800 | ≥ 100,000 | ≤ 50,000 |
| 8 | Jinan/Shandong | ≥ 7,000 | ≤ 1,200 | ≥ 150,000 | ≤ 80,000 |
| 9 | Yantai/Shandong | ≥ 6,000 | ≤ 1,000 | ≥ 120,000 | ≤ 50,000 |
| 10 | Weihai/Shandong | ≥ 10,000 | ≤ 1,500 | ≥ 150,000 | ≤ 80,000 |

In the table 1, we only listed investigation data in ten areas which belong to four different provinces/cities, respectively. In order to integrate investigation data in the same province (or city), we take the average data in the same province (or city) as "integrated data" for each province (or city). Apparently, the more cities are investigated for each province, the more accurate the integrated data will be. To improve the accuracy, we take an "*elasticity value*" for each type integrated data respectively, in which the elasticity value is ± 500/month for "high income", ±100/month for "low income", ± 20,000 for "high deposit", and ± 10,000 for "low deposit". We thus obtained the following "integrated data for each province/city" (Table 2):

**Table 2**. The integrated data in each province/city

| Province | high income (± 500) | low income (±100) | high deposit (± 20,000) | low deposit (± 10,000) |
|---|---|---|---|---|
| Shanghai | ≥ 14,400 | ≤ 2,000 | ≥ 210,000 | ≤ 100,000 |
| Jiangsu | ≥ 11,000 | ≤ 1,340 | ≥ 160,000 | ≤ 82,000 |
| Anhui | ≥ 5,000 | ≤ 920 | ≥ 100,000 | ≤ 56,000 |
| Shandong | ≥ 7,000 | ≤ 1,100 | ≥ 124,000 | ≤ 68,000 |

(2) *Formal representation of the alternative schemes*

Obviously, the attributes of decision-making schemes in the example all have fuzziness. Among them, "*high savings*", "*low savings*", "*moderate savings*", "*high income*", "*low income*" and "*moderate income*" are different fuzzy sets. According to SCOI, these fuzzy sets have the following relationships:

- *low savings* is opposite negation of *high savings*, *moderate savings* is intermediary negation of *low savings* and *high savings*.
- *low income* is opposite negation of *high income*, *moderate income* is intermediary negation of *low income* and *high income*.

Based on SCOI, these different fuzzy sets can be formally represented as follows:

$A$: "high savings"; $A^\neg$: "low savings"; $A^\sim$: "moderate savings"
$B$: "high income"; $B^\neg$: "low income"; $B^\sim$: "moderate income"

where $A^\neg$ and $A^\sim$ are the opposite negation and intermediary negation of $A$, respectively. $B^\neg$ and $B^\sim$ are the opposite negation and intermediary negation of $B$, respectively.

The conclusions for the alternatives can be formally represented as follows:

Bank(*surplus*): the remaining monthly income after living expenses that is deposited into the bank.

Consume(*surplus*): the remaining income after living expenses is used for consumption.

More($a$, $b$): $a$ is greater than $b$.

Since the four alternative schemes are all "*if....then...*" inference form, the four alternative schemes can be formally expressed as:

(a) $A^\neg \to$ Bank(*surplus*)
(b) $A \wedge B \to$ Consume(*surplus*)
(c) $A \wedge B^\sim \to$ More(Consume(*surplus*), Bank(*surplus*))
(d) $A^\sim \wedge B^\sim \to$ More(Bank(*surplus*), Consume(*surplus*))

(3) *Membership degree of data to fuzzy sets*

Due to the fuzzy sets in the alternative schemes determine the attributes of the alternative schemes, so determining the membership degree of the income (or deposit) data in Table 2 to fuzzy sets is the basis for decision-making.

Let the monthly income be $x$, and the deposit be $y$. As can be seen from Table 2, the integrated data has the following characteristics: for four different provinces/cities, if $x \geq 14{,}400$, then the membership degree $B(x)$ of $x$ to the fuzzy set $B$ is 1; if $x \leq 920$, then the membership degree $B^\neg(x)$ of $x$ to the fuzzy set $B^\neg$ is 1. Similarly, if the savings $y \geq 210{,}000$, then the membership degree $A(y)$ of $y$ to the fuzzy set $A$ is 1; if $y \leq 56{,}000$, then the membership degree $A^\neg(y)$ of $y$ to the fuzzy set $A^\neg$ is 1.

Based on this characteristic of the integrated data, we provide a method for calculating the membership degree of data to the fuzzy sets using SCOI, Euclidean Distance $d(t_1, t_2) = |t_1 - t_2|$, and the "Distance Ratio Function" (Zhang, et al, 2009) as follows.

The membership degree $B(x)$ of $x$ to the fuzzy set "high income", and the membership degree $A(y)$ of $y$ to the fuzzy set "high savings":

$$B(x) = \begin{cases} 0, & \text{when } x \leq \alpha_F + \varepsilon_F \\ \dfrac{d(x, \alpha_F + \varepsilon_F)}{d(\alpha_F + \varepsilon_F, \alpha_T - \varepsilon_T)}, & \text{when } \alpha_F + \varepsilon_F < x < \alpha_T - \varepsilon_T \\ 1, & \text{when } x \geq \alpha_T - \varepsilon_T \end{cases} \quad \text{(I)}$$

$$A(y) = \begin{cases} 0, & \text{when } y \leq \alpha_F + \varepsilon_F \\ \dfrac{d(y, \alpha_F + \varepsilon_F)}{d(\alpha_F + \varepsilon_F, \alpha_T - \varepsilon_T)}, & \text{when } \alpha_F + \varepsilon_F < y < \alpha_T - \varepsilon_T \\ 1, & \text{when } y \geq \alpha_T - \varepsilon_T \end{cases} \quad \text{(II)}$$

where $\alpha_T$ is the maximum value of "high income" (or "high deposit") data in the table 2, and $\varepsilon_T$ is the positive elasticity value of $\alpha_T$; $\alpha_F$ is the minimum value of "low income" (or "low deposit") data in the table 2, and $\varepsilon_F$ is the positive elasticity value of $\alpha_F$.

Specifically, in Table 2, for the monthly income data $x$: $\alpha_T = 14{,}400$, $\varepsilon_T = 500$, $\alpha_F = 920$ and $\varepsilon_F = 100$; for the deposit data $y$: $\alpha_T = 210{,}000$, $\varepsilon_T = 20{,}000$, $\alpha_F = 56{,}000$ and $\varepsilon_F = 10{,}000$. Therefore, $B(x)$ and $A(y)$ can be solved as follows:

$$B(x) = \begin{cases} 0, & \text{when } x \leq 1{,}020 \\ \dfrac{d(x, 1{,}020)}{d(1{,}020, 13{,}900)}, & \text{when } 1{,}020 < x < 13{,}900 \\ 1, & \text{when } x \geq 13{,}900 \end{cases} \quad \text{(III)}$$

$$A(y) = \begin{cases} 0, & \text{when } y \leq 66{,}000 \\ \dfrac{d(y, 66{,}000)}{d(66{,}000, 190{,}000)}, & \text{when} 66.000 < y < 190{,}000 \\ 1, & \text{when } y \geq 190{,}000 \end{cases} \quad \text{(IV)}$$

According to the definition of SCOI in Section 4.1, $B^{\lnot}$ is the opposite negation of $B$, and $A^{\lnot}$ is the opposite negation of $A$. So, the membership degree $B(x)$ of $x$ to the fuzzy set "low income" and the membership degree $A(y)$ of $y$ to the fuzzy set "low savings":

$$B^{\lnot}(x) = 1 - B(x), \quad A^{\lnot}(y) = 1 - A(y).$$

Based on (III) and (IV), the membership degree $B^{\sim}(x)$ of $x$ to the fuzzy set "moderate income", the membership degree $A^{\sim}(y)$ of $y$ to the fuzzy set "moderate savings" can be solved by the definition of SCOI.

(4) *"Threshold" of the range of values of the membership degree*

In Section 4.1, we have pointed out that $\lambda\ (\lambda \in (0, 1))$ in the definition of SCOI is a variable parameter introduced in order to determine the intermediary negation set $S^{\sim}$ of a fuzzy set $S$. Since $\lambda$ is variable, the size and variation of $\lambda$ determine the size and range of values for $S(x)$, $S^{\lnot}(x)$ and $S^{\sim}(x)$, i.e. $\lambda$ is a "threshold" for the range of values of these membership degrees (the significance of threshold $\lambda$ is shown in Fig.7). Therefore, to determine whether a monthly income (or deposit) data is high income (or high deposit) or low income (or low deposit), it is necessary to determine the range of values of the membership degree of the monthly income (or deposit) data to the fuzzy sets, and also the relevant threshold. In this regard, we propose a method to determine the threshold $\lambda$ of the range of values of the membership degree as follows:

For the monthly income data $x$, the membership degree $B(x)$ of $x$ to fuzzy set "high income" can be solved by (I). Thus, for the monthly income data 11,000 about 'Jiangsu' in Table 2,

$$B(11{,}000) = \frac{d(11{,}000,\ 1{,}020)}{d(1{,}020,\ 13{,}900)} = 0.775.$$

However, since 11,000 is the minimum value (lower limit) in regard to '*high income*', it should be $B(11{,}000) = 1$.

To eliminate the inconsistency of $B(11{,}000)$ caused by insufficient data and data distortion, we take the average of 0.775 and 1 (i.e., ½ (0.775 + 1) = 0.888) as the "threshold" α, which determines the range of values of the membership degree $B(x)$ of the monthly income data $x$ of 'Jiangsu' to fuzzy set "high income". That is, if $B(x) \geq \alpha$, then it shows that $x$ is '*high income*' in 'Jiangsu'.

Similarly, for other provinces/city in Table 2, we can determine the corresponding thresholds for the range of values of $B(x)$.

For the deposit data $y$, the same method can be used to determine the thresholds of the range of values of the membership degree $A(y)$ of $y$ to the fuzzy set "high savings". All thresholds are shown in the following table (Table 3).

**Table 3**. For each province/city, all thresholds of the range of the membership degrees of the data to the fuzzy sets "high income" and "high savings"

| Provinces/City  Fuzzy sets | Shanghai | Jiangsu | Anhui | Shandong |
|---|---|---|---|---|
| "high income" | 1.000 | 0.888 | 0.655 | 0.732 |
| "high savings" | 1.000 | 0.879 | 0.637 | 0.734 |

In Table 3, we only determined the thresholds of the range of values of the membership degrees of the data to the fuzzy sets "high income" and "high savings" for each province/city. To make the threshold general, take the average of the thresholds in the same row in Table 3 as the thresholds of the range of values of the membership degree $B(x)$ and the range of values of the membership degree $A(y)$, respectively. The resulting general thresholds are shown in the following table (Table 4).

**Table 4**. Thresholds of the range of values of the membership degrees of the data to the fuzzy sets "high income" and "high savings"

| fuzzy sets | "high income" | "high savings" |
|---|---|---|
| membership degree | $B(x)$ | $A(y)$ |
| threshold $\lambda$ | 0.819 | 0.813 |

*The meaning of threshold* $\lambda$: For any monthly income data $x$, if $B(x) \geq \lambda = 0.819$, then $x$ is a '*high income*'; if $B(x) \leq 1-\lambda$, then $x$ is a '*low income*'. For any deposit data $y$, if $A(y) \geq \lambda = 0.813$, then $y$ is a '*high deposit*'; if $A(y) \leq 1-\lambda$, then $y$ is a '*low deposit*'. The meaning of thresholds is shown by the following figure (Fig.13).

According to SCOI, the fuzzy set "low income" is opposite negation of "high income", the fuzzy set "low savings" is opposite negation of "high savings", and $B^\neg(x) = 1-B(x)$ and $A^\neg(y) = 1-A(y)$. Based on the above threshold $\lambda$ of $B(x)$ and $A(y)$ and their meanings, thus there are $B^\neg(x) \geq \lambda$ and $A^\neg(y) \geq \lambda$. Thus, the thresholds for the range of values of the membership degrees $B^\neg(x)$ and $A^\neg(y)$ are shown in the table below (Table 5).

**Table 5**. Thresholds of the range of values of the membership degrees of the data to the fuzzy sets "low income" and "low savings"

| fuzzy sets | "low income" | "low savings" |
|---|---|---|
| membership degree | $B^\neg(x)$ | $A^\neg(y)$ |
| threshold $\lambda$ | 0.819 | 0.813 |

*The meaning of threshold* $\lambda$: For any monthly income data $x$, if $B^\neg(x) \geq \lambda = 0.819$, then $x$ is a '*low income*'. For any deposit data $y$, if $A^\neg(y) \geq \lambda = 0.813$, then $y$ is a '*low deposit*'. The meaning of thresholds is shown by the following figure (Fig.13).

In Table 4, since the threshold λ for the range of values of $B(x)$ is 0.819 > ½, and the threshold λ for the range of values of $A(y)$ is $0.813 > \frac{1}{2}$, according to the definition of SCOI and the proposition 1 in Section 4.1.1, the membership degree $\tilde{B}(x)$ of $x$ to the fuzzy set "moderate income" and the membership degree $\tilde{A}(y)$ of $y$ to the fuzzy set "moderate savings" should satisfy the following relations:

$$\lambda \geq \tilde{B}(x) \geq 1-\lambda, \lambda \geq \tilde{A}(y) \geq 1-\lambda$$

Therefore, the thresholds of the range of values of $\tilde{B}(x)$ is $\lambda_{B1}$ and $\lambda_{B2}$: $\lambda_{B1} = \lambda = 0.819$, $\lambda_{B2} = 1-\lambda = 0.181$; the thresholds of the range of values of $\tilde{A}(y)$ is $\lambda_{A1}$ and $\lambda_{A2}$: $\lambda_{A1} = \lambda = 0.813$, $\lambda_{A2} = 1-\lambda = 0.187$. As shown in the following table (Table 6).

**Table 6**. Thresholds of the range of values of the membership degrees of the data to the fuzzy sets "moderate income" and "moderate savings"

| fuzzy sets | "moderate income" | "moderate savings" |
|---|---|---|
| membership degree | $\tilde{B}(x)$ | $\tilde{A}(y)$ |
| threshold | $\lambda_{B1}, \lambda_{B2}$ | $\lambda_{A1}, \lambda_{A2}$ |

*The meaning of thresholds*: For any monthly income data $x$, if $\lambda_{B1} \geq \tilde{B}(x) \geq \lambda_{B2}$, then $x$ is a '*middle income*'. For any savings data $y$, if $\lambda_{A1} \geq \tilde{A}(y) \geq \lambda_{A2}$, then $y$ is a '*middle deposit*'. The meaning of thresholds is shown in the following figure (Fig. 13).

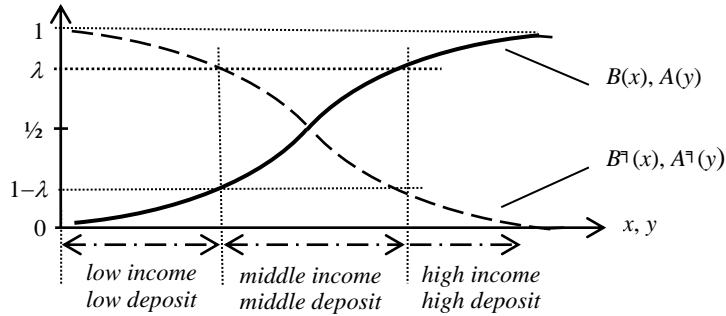

**Fig. 13**. The thresholds of the range of values of the membership degrees $B(x)$, $A(y)$, $B^⇁(x)$ and $A^⇁(y)$, and their meanings

(5) *Inference*

In the example, all four alternatives follow the inference form of "*if... then...*". Therefore, we use "fuzzy production rules" (Dung, et al, 2002) to make reasoning decisions for the example. The general form of fuzzy production rules is as follows:

$$S_1, S_2, \ldots, S_m \to Q \,|\, \langle bd, (\alpha_1, \alpha_2, \ldots, \alpha_m) \rangle \qquad (V)$$

where $S_i$ ($i = 1, 2, \ldots, m$) are fuzzy sets, which denoted the premises of the rule, $Q$ denoted the conclusions of inference. $bd$ ($0 \leq bd \leq 1$) denoted the belief degree of rule, $\alpha_i$ ($\alpha_i \in [0, 1]$) denoted the threshold of the range of values of the membership degree $S_i(x)$ of $x$ to $S_i$.

The meaning of the fuzzy production rule (V) is as follows:

"If each $S_i(x) \geq \alpha_i$, then deduce $Q$ from $S_1, S_2, \ldots, S_m$ with belief degree $bd$." (VI)

In the example, since of the individual M's monthly income is 5,000 and the bank deposit is 120,000, according to (I) and (II), the membership degree $B(5,000)$ of 5,000 to the fuzzy set "high income", and the membership degree $A(120,000)$ of 120,000 to the fuzzy set "high savings":

$$B(5,000) = \frac{d(5,000, 1,020)}{d(1,020, 13,900)} = 0.309, \quad A(120,000) = \frac{d(120,000, 66,000)}{d(66,000, 190,000)} = 0.435.$$

By the definition of SCOI, the membership degree $A^⇁(120,000)$ of 120,000 to the fuzzy set "low savings":

$$A^⇁(120,000) = 1 - A(120,000) = 0.565.$$

Based on the above, we can infer the following:

1) For the alternative scheme (a) in the example, the fuzzy set in (a) has "low savings". The threshold $\lambda$

= 0.813 of the range of values of the membership degree $A^{\neg}(y)$ by the table 5. Thus, according to (V),
$$A^{\neg}(y) \to Bank(surplus) \mid \langle bd, (0.813) \rangle$$
Because of $A^{\neg}(120,000) = 0.565 < 0.813$, $A^{\neg}(120,000)$ does not satisfy (VI). Therefore, the alternative scheme (a) cannot be adopted.

2) For the alternative scheme (b) in the example, the fuzzy sets in (b) have "high savings" and "high income". The threshold $\lambda = 0.813$ of the range of values of the membership degree $A(y)$ and the threshold $\lambda = 0.819$ of the range of values of the membership degree $B(x)$ by the table 4. Thus, according to (V),
$$A(y), B(x) \to Consume(surplus) \mid \langle bd, (0.813, 0.819) \rangle$$
Because of $A(120,000) = 0.435 < 0.813$ and $B(5,000) = 0.309 < 0.819$, $A(120,000)$ and $B(5,000)$ does not satisfy (VI). Therefore, the alternative scheme (b) cannot be adopted.

3) For the alternative scheme (c) in the example, the fuzzy sets in (c) have "high savings" and "moderate income". The threshold $\lambda = 0.813$ of the range of values of the membership degree $A(y)$ by the table 4. The minimal threshold $\lambda_B = 0.181$ of the range of values of the membership degree $B\tilde{}(x)$ by the table 6. Thus, according to (V),
$$A(y), B\tilde{}(x) \to More(Consume(surplus), Bank(surplus)) \mid \langle bd, (0.813, 0.181) \rangle$$
Because of $A(120,000) = 0.435 < 0.813$, $A(120,000)$ does not satisfy (VI). Therefore, the alternative scheme (c) cannot be adopted.

4) For the alternative scheme (d) in the example, the fuzzy sets in (d) have "moderate savings" and "moderate income". The minimal threshold $\lambda_{A2}$ of the range of values of the membership degree $A\tilde{}(y)$ is 0.187, and the minimal threshold $\lambda_{B2}$ of the range of values of the membership degree $B\tilde{}(x)$ is 0.181 by the table 6. Thus, according to (V),
$$A\tilde{}(y), B\tilde{}(x) \to More(Bank(surplus), Consume(surplus)) \mid \langle bd, (0.187, 0.181) \rangle$$
Due to the threshold $\lambda = 0.813$ of the range of values of the membership degree $A(y)$ in the table 4, $\lambda > ½$ and $1-\lambda < A(120,000) = 0.435 < \lambda$, so $A\tilde{}(120,000)$ is the case (e) in the definition of SCOI (Definition 1 in Section 4.1). That is, $A\tilde{}(120,000) = A(120,000)$. Thus, $A\tilde{}(120,000) > 0.187$. Similarly, due to the threshold $\lambda = 0.819$ of the range of values of the membership degree $B(x)$ in the table 4, $\lambda > ½$ and $1-\lambda < B(5,000) = 0.309 < \lambda$, $B\tilde{}(5,000) = B(5,000)$ according to the case (e) in the definition of SCOI. Thereby $B\tilde{}(5,000) > 0.181$. Thus, $A\tilde{}(120,000)$ and $B\tilde{}(5,000)$ satisfy (VI). Therefore, the alternative scheme (d) can be adopted.

(6) *Conclusions of decision-making*

For the individual M in the example, the alternative scheme (a), (b) and (c) are not available, (d) is the most suitable decision- making scheme for M.

The above indicates that using SCOI and LCOI to handle knowledge with different types of negativity is effective.

## 6. Conclusion and future work

In various types of knowledge, negative knowledge plays an important role. Therefore, how to understand, distinguish, express, and compute negative knowledge is a fundamental issue in the research and processing of knowledge. In this paper, we analyze and examine the understanding and characteristics of "negation" in various fields of knowledge, such as philosophy, logic, and linguistics. Based on the distinction between the concepts of "contradiction" and "opposition", we propose from a conceptual perspective that there are three different types of negation present in knowledge: "contradictory negation", "opposite negation", and "intermediary negation".

To establish a mathematical foundation that can fully reflect the internal relations, properties and laws of these different negations, we propose a "SCOI: sets with contradictory negation, opposite negation and intermediary negation" and "LCOI: logic with contradictory negation, opposite negation and intermediary negation". In addition, we also discussed the main operational properties of SCOI and the formal inference relations in LCOI, proving that LCOI is both sound and complete under the given three-valued semantic interpretation of LCOI.

The objectivity and expressive power of "intermediary negation" in SCOI and LCOI are demonstrated through examples. SCOI and LCOI can express statements of the type '$x$ is neither a positive nor a negative integer' and 'he is neither rich nor poor'. For statements with more complex negativity, whether they are clear or fuzzy, they can be expressed by the "intermediary negation" of a set in SCOI and the "intermediary negation" of an atomic formula in LCOI.

In order to demonstrate that SCOI and LCOI are suitable for representing and reasoning about knowledge with different forms of negation, we will apply SCOI to a multi-attribute decision-making example, showing that both SCOI and LCOI are effective in knowledge processing.

Based on the results of this paper, we will further study and improve the theories of SCOI and LCOI. We will mainly study the decomposition theorem, cut set and closeness degree of SCOI, and the predicate logic calculus and its semantics based on LCOI. On this basis, we will continue to study the applications of SCOI and LCOI in typical knowledge processing fields.